\documentclass[a4paper,fleqn]{cas-dc}

\usepackage[numbers]{natbib}

\usepackage{algorithm}
\usepackage[noend]{algorithmic}

\usepackage{threeparttable}
\usepackage{multirow}
\usepackage[switch, pagewise]{lineno}
\usepackage{stfloats} 
\usepackage{color}
\usepackage{tabularx}
\usepackage{booktabs}
\usepackage{graphicx} 
\usepackage{hyperref}
\usepackage{amsmath}
\usepackage{amssymb}
\usepackage{dsfont}
\usepackage{float} 
\usepackage{subfig} 
\usepackage{graphicx, subcaption}

\usepackage{pifont} 
\newcommand{\cmark}{\ding{51}} 
\newcommand{\xmark}{\ding{55}}

\usepackage{algorithmic}
\usepackage{algorithm}

\def\tsc#1{\csdef{#1}{\textsc{\lowercase{#1}}\xspace}}
\tsc{WGM}
\tsc{QE}
\tsc{EP}
\tsc{PMS}
\tsc{BEC}
\tsc{DE}

\begin{document}
\let\WriteBookmarks\relax
\def\floatpagepagefraction{1}
\def\textpagefraction{.001}

\shorttitle{VID-AD: A Dataset for Image-Level Logical Anomaly Detection under Vision-Induced Distraction}

\shortauthors{Hiroto Nakata et~al.}

\title [mode = title]{VID-AD: A Dataset for Image-Level Logical Anomaly Detection under Vision-Induced Distraction}                      

%
\author[1]{Hiroto Nakata}[type=editor]



\ead{nkt.hiroto.research@gmail.com}



\affiliation[1]{organization={Department of Engineering, University of Fukui},
    addressline={3-9-1 Bunkyo},
    city={Fukui-city},
    postcode={910-8507},
    country={Japan}}

\author[2]{Yawen Zou}[type=editor]
\ead{yawenzou93@gmail.com}
\affiliation[2]{organization={Graduate School of Science and Engineering, University of Toyama},
    addressline={3190 Gofuku},
    city={Toyama-city},
    postcode={930-8555},
    country={Japan}} 
    
\author[1]{Shunsuke Sakai}[type=editor]
\ead{sshunsuke0102@gmail.com}

\author[1]{Shun Maeda}[type=editor]
\ead{s-maeda@monju.fuis.u-fukui.ac.jp}

\author[3]{Chunzhi Gu}[type=editor]
\ead{czgu@ieee.org}
\affiliation[3]{organization={Faculty of Engineering, University of Fukui},
    addressline={3-9-1 Bunkyo},
    city={Fukui-city},
    postcode={910-8507},
    country={Japan}}

\author[2]{Yijin Wei}[type=editor]
\ead{weiyijin0816@gmail.com}

\author[4]{Shangce Gao}[type=editor]
\ead{gaosc@eng.u-toyama.ac.jp}
\affiliation[4]{organization={Faculty of Engineering, University of Toyama},
   addressline={3190 Gofuku},
   city={Toyama-city},
   postcode={930-8555},
   country={Japan}} 

\author[4]{Chao Zhang}[type=editor]
\cormark[1]
\ead{zhang@eng.u-toyama.ac.jp}


\cortext[cor1]{Corresponding author}



\begin{abstract}
Logical anomaly detection in industrial inspection remains challenging due to variations in visual appearance (e.g., background clutter, illumination shift, and blur), which often distract vision-centric detectors from identifying rule-level violations. However, 
existing benchmarks rarely provide controlled settings where logical states are fixed while such nuisance factors vary.
To address this gap, we introduce VID-AD, a dataset for logical anomaly detection under vision-induced distraction. It comprises 10 manufacturing scenarios and five capture conditions, totaling 50 one-class tasks and 10,395 images.
Each scenario is defined by two logical constraints selected from quantity, length, type, placement, and relation, with anomalies including both single-constraint and combined violations. We further propose a language-based anomaly detection framework that relies solely on text descriptions generated from normal images.
Using contrastive learning with positive texts and contradiction-based negative texts synthesized from these descriptions, our method learns embeddings that capture logical attributes rather than low-level features.
Extensive experiments demonstrate consistent improvements over baselines across the evaluated settings.
The dataset is available at: \url{https://github.com/nkthiroto/VID-AD}.
\end{abstract}



\maketitle

\section{Introduction}
\label{sec:introduction}

Anomaly detection is a fundamental problem for ensuring safety and reliability in a wide range of applications, including medical diagnosis \cite{bercea2024towards, bercea2025evaluating}, autonomous driving \cite{shoeb2025out, ren2025efficient}, and industrial inspection \cite{li2025survey, shukla2025systematic}. 
In industrial visual inspection, a central challenge is that anomalies are not always manifested as obvious structural defects, such as scratches, dents, or contamination. 
In many cases, defects are defined by violations of global logical constraints, including differences in quantity, length, type, placement, and inter-object relations. 
Such logical anomalies \cite{bergmann2022beyond} can be visually subtle, and their discriminative cues are often less pronounced and more spatially dispersed than those of localized defects.
\begin{figure}[pos=tb]
\centering
\includegraphics[width=0.45\textwidth]{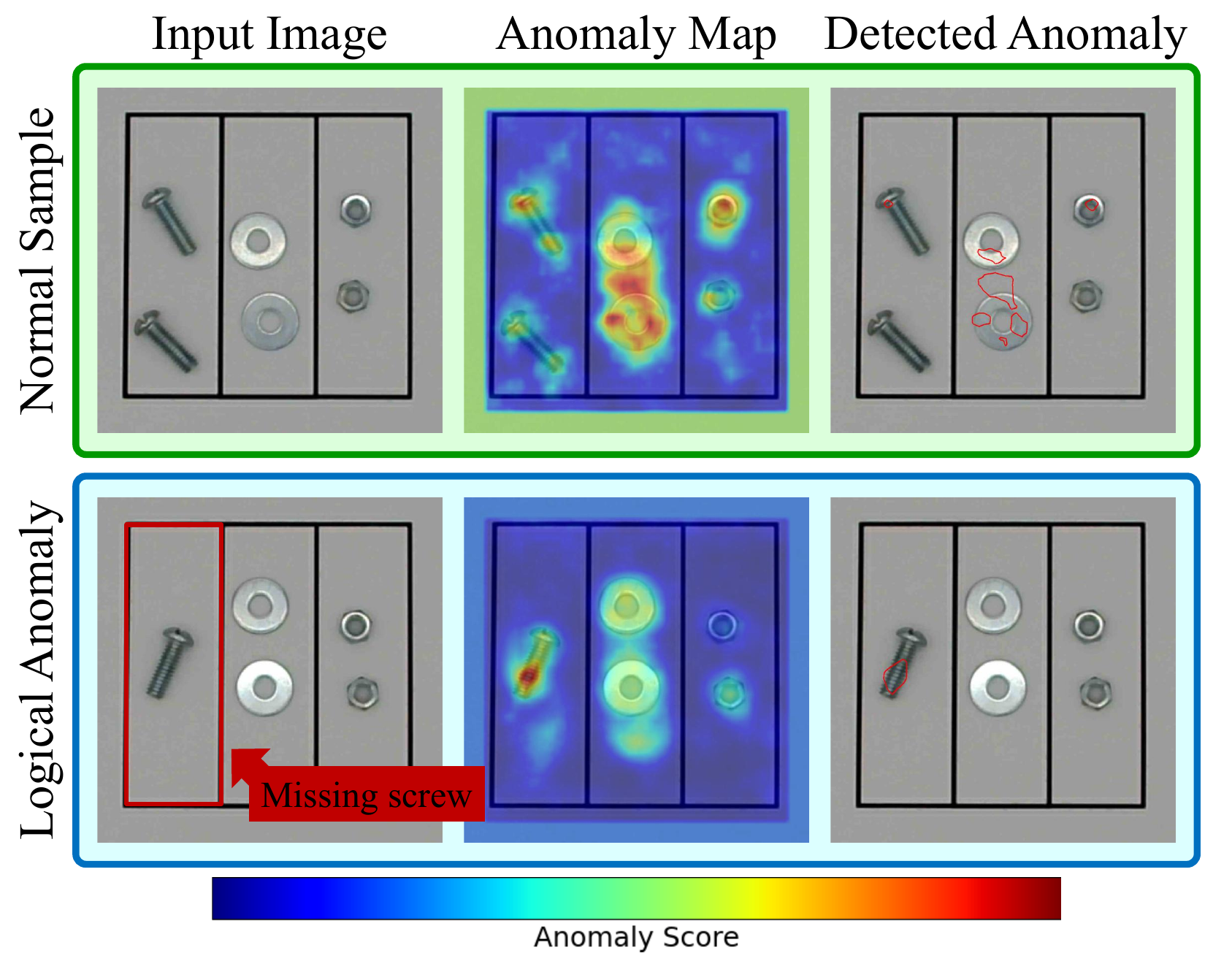}
\caption{
\textbf{Example results} from the \textit{Tools} scenario in VID-AD.
The top row (green border) shows a Normal Sample, and the bottom row (blue border) shows a Logical Anomaly where a screw is missing.
From left to right, we present the Input Image, Anomaly Map, and Detected Anomaly, where red highlighted areas indicate regions classified as anomalous based on a detection threshold.
EfficientAD \cite{batzner2024efficientad} produces strong anomaly responses even for the Normal Sample, resulting in false positives; it also fails to clearly localize the missing screw in the Logical Anomaly case.
}
\label{ex_Tools}
\end{figure}

Most existing unsupervised anomaly detection methods \cite{cohen2020sub, defard2021padim, roth2022towards, an2015variational, vasilev2020q, creswell2018generative, schlegl2019f, batzner2024efficientad, hsieh2024csad} for visual inspection rely on local visual cues and typically employ patch-centric representations. While effective for structural anomalies, these methods inherently struggle to model the global consistency required for identifying logical constraint violations. This limitation stems from the fact that real inspection environments exhibit low-level visual variations such as background changes, illumination shifts, and blurs. Even when the underlying logical state is unchanged, such variations can distract vision-based detectors and lead to false positives. As illustrated in Fig. \ref{ex_Tools}, a state-of-the-art detector such as EfficientAD \cite{batzner2024efficientad} produces strong anomaly responses even for normal samples under such distractions, resulting in false positives while failing to clearly localize the actual logical anomaly. This failure demonstrates that patch-centric visual representations generate spurious responses for logical violations under environmental noise.

\begin{figure*}[pos=t]
\centering
\includegraphics[width=1.0\textwidth]{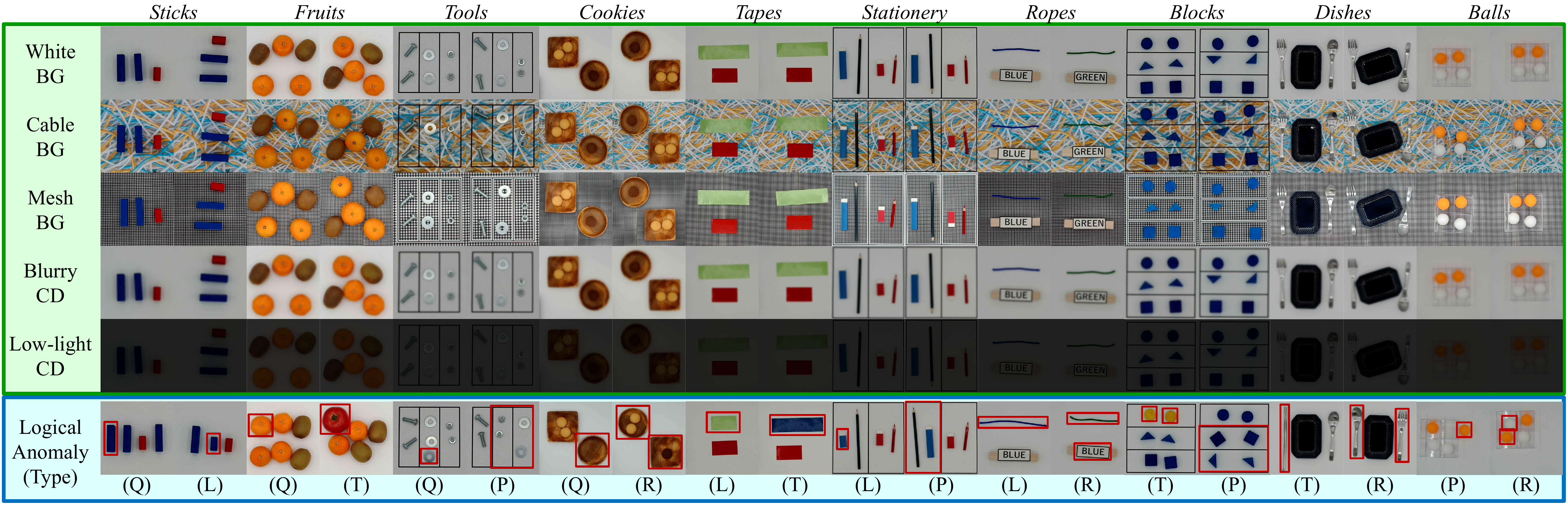}
\caption{
\textbf{Overview of VID-AD.}
Columns correspond to the 10 manufacturing scenarios (\textit{Sticks}, \textit{Fruits}, \textit{Tools}, \textit{Cookies}, \textit{Tapes}, \textit{Stationery}, \textit{Ropes}, \textit{Blocks}, \textit{Dishes}, \textit{Balls}), with two representative images shown for each scenario.
The first five rows correspond to the five capture conditions: White BG (default), Cable BG, Mesh BG, Blurry CD, and Low-light CD (BG: background, CD: condition).
The bottom row shows logical anomaly examples for each scenario, with red boxes highlighting the anomalous regions.
Each scenario is defined by a pair of logical constraints, as indicated beneath each scenario name (abbreviations: Q = Quantity, L = Length, T = Type, P = Placement, R = Relation).
Each cell in the first five rows represents a one-class benchmark task defined by a scenario and a capture condition, yielding 50 tasks in total (10 scenarios $\times$ five capture conditions).
}
\label{introduction}
\end{figure*}

However, current benchmarks seldom incorporate the controlled environmental variations necessary to assess the robustness of logical anomaly detection against visual distractions.
Widely used datasets \cite{bergmann2019mvtec, mishra2021vt, zou2022spot, bergmann2022beyond, zhang2024learning} primarily focus on structural anomalies, leaving logical violations, particularly those occurring under varying capture conditions, largely underexplored. To bridge these gaps, we introduce VID-AD, a dataset specifically designed to evaluate robustness across low-level visual variations while preserving well-defined logical constraints. VID-AD contains 10 manufacturing scenarios and five capture conditions (50 one-class tasks; 10,395 images), as illustrated in Fig. \ref{introduction}. For each scenario, anomalies are defined by two logical constraints, including single-constraint and combined violations. By design, each capture condition changes only visual appearance while keeping the underlying logical constraints fixed, enabling controlled evaluation of robustness to vision-induced distractions.

Beyond the dataset, we propose a text-based anomaly detection framework that trains a natural language processing (NLP) model \cite{vaswani2017attention, devlin2019bert} solely on language representations derived from normal images, without learning visual feature embeddings \cite{lee2024nv, wang2024improving}. 
Specifically, a Vision-Language Model (VLM), often implemented as a large multimodal language model \cite{yin2024survey}, is used to convert normal images into structured textual descriptions using a text prompt. To enable effective learning without access to real anomalous images, we introduce a text-only rewriting strategy to synthesize negative text examples  from these normal descriptions. Specifically, our method performs replacement-only edits to enforce attribute-level contradictions in properties such as quantity, relation, or type, while strictly preserving the original text structure.
We then perform contrastive learning \cite{chen2020simple, radford2021learning} with positive text examples and semantically perturbed negative text examples to suppress irrelevant visual cues while preserving global structural semantics. 
Extensive experiments on VID-AD demonstrate that representative vision-based baselines degrade under the proposed setting, whereas our method consistently improves performance across all capture conditions.

In summary, our contributions are threefold:

\begin{itemize}
    \item We propose VID-AD, a new benchmark dataset that exclusively introduces vision-induced distractions to support the evaluation of logical anomalies in industrial inspection.
    \item We develop a text-based framework that learns logical consistency through contrastive learning with constrained rewritten normal descriptions, bypassing the need for real anomalous training images. 
    \item Extensive experiments on VID-AD demonstrate that our method consistently outperforms existing vision-based baselines across all capture conditions, achieving state-of-the-art performance.
\end{itemize}

\section{Related Work}
\label{sec: related work}

We first review existing datasets for anomaly detection in visual inspection and highlight their limitations, thereby motivating the necessity of our proposed VID-AD dataset. 
We then discuss prior methods for anomaly detection in visual inspection and contrast our approach with them, emphasizing how it addresses their shortcomings.

\subsection{VAD Datasets}
Most existing datasets for unsupervised anomaly detection in industrial visual inspection focus on structural defects, such as scratches, dents, contamination, or surface damage.
Representative datasets include MVTec AD \cite{bergmann2019mvtec}, BTAD \cite{mishra2021vt}, and VisA \cite{zou2022spot}, which collectively cover diverse structural defect types across a wide range of industrial categories and have played a critical role in advancing visual anomaly detection.
In addition to these specialized datasets, recent studies have adapted general-purpose data to establish VAD benchmarks, such as COCO-AD \cite{zhang2024learning} built upon COCO \cite{lin2014microsoft}.
However, the anomalies in these datasets are largely characterized by visually explicit cues at the pixel or texture level, which renders them inadequate for systematically studying logical anomalies that arise from violations of global consistency constraints.

Beyond these standard benchmarks, recent datasets extend industrial anomaly detection toward more practical acquisition settings and evaluation axes.
For example, several benchmarks feature multi-view capture or diverse viewpoints to assess generalization in realistic inspection environments (PAD \cite{zhou2023pad}, Real-IAD \cite{wang2024real}, MANTA \cite{fan2025manta} and CableInspect-AD \cite{arodi2024cableinspect}).
Meanwhile, benchmarks such as MVTec AD 2 \cite{heckler2025mvtec} and AutVI \cite{carvalho2024detecting} encompass a wider range of industrial scenarios with more challenging imaging conditions to better capture real-world variability.
In addition, robustness-oriented benchmarks explicitly evaluate performance under imaging noise such as uneven illumination and blur (RAD \cite{cheng2024rad}, Robust AD \cite{pemula2025robust}, PIAD \cite{yang2025piad}).
These efforts substantially improve robustness to real-world variability and acquisition shifts, but they primarily focus on visual anomalies while overlooking logical anomalies under controlled vision-induced distractions.

A notable exception is MVTec LOCO AD \cite{bergmann2022beyond}, which is the first to systematically incorporate logical anomalies beyond purely structural defects.
While this is an important step toward logic-aware evaluation, it lacks a dedicated setting where low-level visual variations are introduced as distractors while the logical state remains unchanged.
As a result, a testbed that jointly targets logical constraint violations and controlled vision-induced distraction remains scarce.
To fill this gap, we introduce the Vision-Induced Distraction Anomaly Dataset (VID-AD).

\subsection{VAD Methods}
Recent unsupervised anomaly detection methods for industrial visual inspection primarily identify anomalies based on visual representations, which can be categorized into several main approaches: patch-level feature deviation, reconstruction discrepancy, representation distillation, component-wise consistency, and recently, training-free few-shot matching. 
Representative instances include PaDiM \cite{defard2021padim} and PatchCore \cite{roth2022towards} (feature deviation on ImageNet-pretrained backbones \cite{krizhevsky2012imagenet}), AnoGAN \cite{schlegl2017unsupervised} and VAE-based methods \cite{kingma2013auto, an2015variational} (reconstruction and latent discrepancies), EfficientAD \cite{batzner2024efficientad} (student–teacher distillation), CSAD \cite{hsieh2024csad} (component-level consistency), and UniVAD \cite{gu2025univad} (training-free few-shot unified detection). 
Collectively, these methods excel at identifying structurally explicit defects that manifest as local visual irregularities, such as scratches, dents, or contamination.

However, their heavy reliance on local visual patterns poses two fundamental challenges to effectively identifying logical anomalies in distracted environments.
First, logical anomalies often yield dispersed or ambiguous visual evidence that requires global consistency reasoning.
Since patch-centric scoring, reconstruction cues, and feature discrepancies focus primarily on local visual patterns, they may fail to capture semantic rule violations that span multiple objects or regions, even when local textures appear normal.
Second, robustness under low-level visual variations remains challenging \cite{taori2020measuring}.
These low-level visual variations include background changes, illumination shifts, and blur, which can alter visual statistics \cite{hendrycks2019benchmarking} without changing the underlying logical state, yet they directly affect patch embeddings, reconstruction fidelity, feature discrepancies, and even component segmentation or matching.
Consequently, vision-centric pipelines can be distracted by nuisance factors, producing false positives on normal samples or unstable evidence for truly anomalous ones under vision-induced distraction.

To overcome these limitations, we propose a text-based anomaly detection framework that avoids learning visual feature embeddings and instead converts each image into a logic-focused description. 
By modeling semantic consistency in the language embedding space, our approach targets logical rule violations beyond purely visual cues and is inherently robust to appearance-induced distractions.

\section{VID-AD Dataset}
\label{sec:dataset}

\begin{table*}[t]
\caption{
\textbf{Comparison} of VID-AD with representative anomaly detection datasets for industrial visual inspection.
Tasks denotes the number of independent one-class benchmark tasks.
VID indicates whether the dataset systematically introduces controlled low-level visual variations (background, illumination, blur) as separate evaluation settings while preserving the underlying logical or structural state.
\cmark: Satisfied.
\xmark: Unsatisfied.
}
\label{tab:dataset-comparison}
\centering
\begin{tabular}{@{}@{\extracolsep{\fill}}lc ccc c cc@{}}
\hline
\multirow{2}{*}{Dataset} & \multirow{2}{*}{Tasks} & \multicolumn{3}{c}{Image Number} & \multirow{2}{*}{VID} & \multicolumn{2}{c}{Anomaly Type} \\ \cline{3-5} \cline{7-8} 
 &  & Normal & Anomaly & All &  & Structure Anomaly & Logical Anomaly \\ \hline
MVTec AD {\cite{bergmann2019mvtec}} & 15 & 4,096 & 1,258 & 5,354 & \xmark & \cmark & \xmark \\
VisA {\cite{zou2022spot}} & 12 & 9,621 & 1,200 & 10,821 & \xmark & \cmark & \xmark \\
Real-IAD {\cite{wang2024real}} & 30 & 99,721 & 51,329 & 151,050 & \xmark & \cmark & \xmark \\
RAD {\cite{cheng2024rad}} & 4 & 1,144 & 1,224 & 2,368 & \cmark & \cmark & \xmark \\
MVTec LOCO AD {\cite{bergmann2022beyond}} & 5 & 2,347 & 993 & 3,340 & \xmark & \cmark & \cmark \\ \hline
VID-AD (Ours) & 50 & 5,000 & 5,395 & 10,395 & \cmark & \xmark & \cmark \\ \hline
\end{tabular}
\end{table*}

While existing benchmarks have significantly advanced industrial anomaly detection, they primarily focus on structural defects \cite{bergmann2019mvtec, mishra2021vt, zou2022spot}. Moreover, even datasets featuring logical anomalies \cite{bergmann2022beyond} often fail to decouple logical violations from visual appearance variations. Consequently, it remains difficult to evaluate a model's true semantic consistency under vision-induced distractions, such as illumination changes or blur.
To address these limitations, we introduce the \textbf{V}ision-\textbf{I}nduced \textbf{D}istraction \textbf{A}nomaly \textbf{D}etection (VID-AD) dataset, which enables controlled evaluation of logical anomaly detection under low-level visual variations.

\subsection{Dataset Overview}
VID-AD is a comprehensive one-class benchmark designed for logical anomaly detection under controlled environmental perturbations. The dataset consists of 10 manufacturing scenarios and five capture conditions, resulting in 50 independent tasks where models are trained exclusively on normal samples. Through this multi-scenario and multi-condition framework, VID-AD provides a more rigorous and systematic evaluation of model stability than previous benchmarks. The core design principle of VID-AD is the decoupling of logical configurations from visual appearance. In real manufacturing inspection, the same logically correct assembly can be captured under substantially different visual conditions, such as background clutter, illumination changes, and defocus.
These low-level visual variations can dominate visual evidence and mislead vision-centric detectors even when the underlying logical state remains unchanged. To address this, VID-AD maintains fixed logical rules within each scenario while systematically varying only the capture conditions.

VID-AD distinguishes itself from previous works through its exclusive focus on logical anomalies and its systematic environmental control, as summarized in Tab.~\ref{tab:dataset-comparison}. While the majority of industrial benchmarks target only structural defects \cite{bergmann2019mvtec, mishra2021vt}, MVTec LOCO AD \cite{bergmann2022beyond} introduces logical anomalies but mixes them with structural ones. By evaluating logical consistency under five diverse capture conditions, our dataset ensures that detection results stem from robust logical understanding instead of a simple reaction to pixel-level perturbations.

\subsection{Scenarios and Logical Anomaly Taxonomy}
We define five types of logical anomalies based on the violation of specific constraints: Quantity, Length, Type, Placement, and Relation. Quantity anomalies specify whether the number of target objects matches the expected count. Length anomalies arise from violations of prescribed length attributes, where logical consistency is evaluated through relative comparisons between instances or against a reference object. Type anomalies occur when an object is replaced by an incorrect category through either fixed or variable substitution. Placement anomalies involve objects appearing outside their designated regions or slots, with layouts that vary across scenarios. Finally, Relation anomalies differ from Placement by focusing on the logical relationships between two or more objects. These anomalies involve violations of rules such as relative spatial positioning, valid pairings, or dependency constraints where the visual attributes of one component must be consistent with the attributes of another.
For example, \textit{Dishes} specifies the left-to-right ordering of fork–plate–spoon (relative position), \textit{Cookies} constrains cookie type and count conditioned on the dish shape (pairing), and \textit{Ropes} requires consistency between a textual label and rope color together with a length constraint defined relative to a reference stick (dependency).

Based on these five aspects, we define ten manufacturing scenarios in VID-AD by pairing two constraints per scenario. Under this design, a normal sample must satisfy both constraints simultaneously, while an anomalous sample violates at least one of them. The paired aspects and corresponding rules for all scenarios are summarized in Tab. \ref{tab:scenarios}. For each scenario, we further construct anomaly subsets that separately isolate violations of the first aspect, the second aspect, or both, which enables a detailed analysis of specific logical violation types.

\begin{table*}[ht]
\caption{
\textbf{Ten scenarios in VID-AD.}
Each scenario is defined by a pair of logical aspects and summarized with a one-line normal rule and representative anomaly patterns. 
Abbreviations: Q = Quantity, L = Length, T = Type, P = Placement, R = Relation; a pair such as Q+L indicates that anomalies are defined by violating either (or both) of the two aspects.
}
\label{tab:scenarios}
\centering
\resizebox{\linewidth}{!}{
\begin{tabularx}{\hsize}{@{}@{\extracolsep{\fill}}l l X l@{}}
\hline
Scenario & Aspects & Normal rule (one line) & Typical anomaly patterns (keywords) \\
\hline
\textit{Sticks} & Q+L &
Two long blue sticks and one short red stick are present. &
missing/extra; long$\leftrightarrow$short \\

\textit{Fruits} & Q+T &
Three oranges and two kiwifruits are present. &
missing/extra; wrong fruit type \\

\textit{Tools} & Q+P &
Two bolts, two washers, and two nuts are present and placed in the Left/Middle/Right bins, respectively. &
missing/extra; wrong bin \\

\textit{Cookies} & Q+R &
Two yellow cookies are on the square dish and one black cookie is on the round dish. &
missing/extra; wrong dish--cookie pairing \\

\textit{Tapes} & L+T &
A long green tape and a short red tape are present (length judged by the relative length of the two tapes). &
long/short/similar mismatch; wrong color type \\

\textit{Stationery} & L+P &
Left bin contains a long black pencil and a long blue eraser; right bin contains a short red pencil and a short red eraser, with the eraser placed left of the pencil within each bin. &
length mismatch; wrong bin/order \\

\textit{Ropes} & L+R &
The rope length is similar to the reference stick, and the rope color matches the text label. &
length mismatch; label--color mismatch \\

\textit{Blocks} & T+P &
Two circle blocks, two triangle blocks, and two square blocks are placed in the Top/Middle/Bottom bins, respectively. &
wrong shape type; wrong bin \\

\textit{Dishes} & T+R &
A fork, a plate, and a spoon are arranged in this left-to-right order. &
wrong utensil type; wrong order \\

\textit{Balls} & P+R &
Orange balls are in the top compartments and white balls are in the bottom compartments of the 2$\times$2 case. &
wrong compartment; color-to-compartment mismatch \\
\hline
\end{tabularx}}
\end{table*}

\subsection{Capture Conditions for Vision-Induced Distraction}
To evaluate model robustness under vision-induced distraction, we define five capture conditions that commonly arise in real inspection environments: White BG (default), Cable BG, Mesh BG, Low-light CD, and Blurry CD. These distractions encompass three major sources of low-level visual variations: background changes (Cable BG, Mesh BG), illumination shifts (Low-light CD), and lens-related degradation (Blurry CD).
Across these conditions, variations are restricted to the background, illumination, or defocus, while scenario rules and logical configurations remain unchanged. This ensures that differences in detection behavior can be attributed to vision-induced distraction rather than changes in the logical state.
Notably, these acquisition variations are treated as separate benchmark tasks rather than augmentations.

\subsection{Benchmark Protocol and Dataset Statistics}
The aforementioned scenarios and capture conditions yield 50 independent tasks. Each task adopts a one-class protocol where training is restricted to normal images, while the test set contains both normal and anomalous samples. A typical task provides 50 training normal images, 50 testing normal images, and approximately 110 testing anomalous images. Collectively, VID-AD comprises 10,395 samples, including 2,500 for training and 7,895 for testing. All images are provided in 1080x1080 JPG format.

The benchmark provides binary labels for normal and anomalous samples along with metadata identifying violations in Quantity, Length, Type, Placement, or Relation. Performance results are reported at the scenario level by aggregating data within each capture condition. Detailed scenario-level statistics, including the specific sample counts for each split and the distribution of logical violation types, are summarized in Tab. \ref{tab:dataset_detail}.

\begin{table*}[ht]
\caption{
\textbf{Per-scenario statistics} of VID-AD reported per capture condition (i.e., per one-class task). 
For each scenario, we report the one-class split (train normal, test normal, test anomaly) and the anomaly breakdown into single-aspect and dual-aspect violations. 
Single-A and Single-B denote violations of the first and second aspect in the scenario definition, respectively (e.g., for Q+L, Single-A indicates Quantity and Single-B indicates Length).}
\label{tab:dataset_detail}
\centering
\begin{tabular}{llcccccc}
\hline
Scenario & Aspects & \# Train Normal & \# Test Normal & \# Test Anomaly & \# Single-A & \# Single-B & \# Dual \\
\hline
\textit{Sticks}      & Q+L & 50 & 50 & 104 & 48 & 48 & 8 \\
\textit{Fruits}      & Q+T & 50 & 50 & 100 & 48 & 44 & 8 \\
\textit{Tools}       & Q+P & 50 & 50 & 110 & 52 & 50 & 8 \\
\textit{Cookies}     & Q+R & 50 & 50 & 106 & 50 & 50 & 6 \\
\textit{Tapes}       & L+T & 50 & 50 & 110 & 50 & 50 & 10 \\
\textit{Stationery}  & L+P & 50 & 50 & 110 & 50 & 50 & 10 \\
\textit{Ropes}       & L+R & 50 & 50 & 110 & 48 & 50 & 12 \\
\textit{Blocks}      & T+P & 50 & 50 & 110 & 52 & 50 & 8 \\
\textit{Dishes}      & T+R & 50 & 50 & 111 & 48 & 48 & 15 \\
\textit{Balls}       & P+R & 50 & 50 & 108 & 48 & 48 & 12 \\
\hline
\end{tabular}
\end{table*}

\section{Proposed Method}
\label{sec:method}

\begin{figure*}[pos=tb]
\centering
\includegraphics[width=1\textwidth]{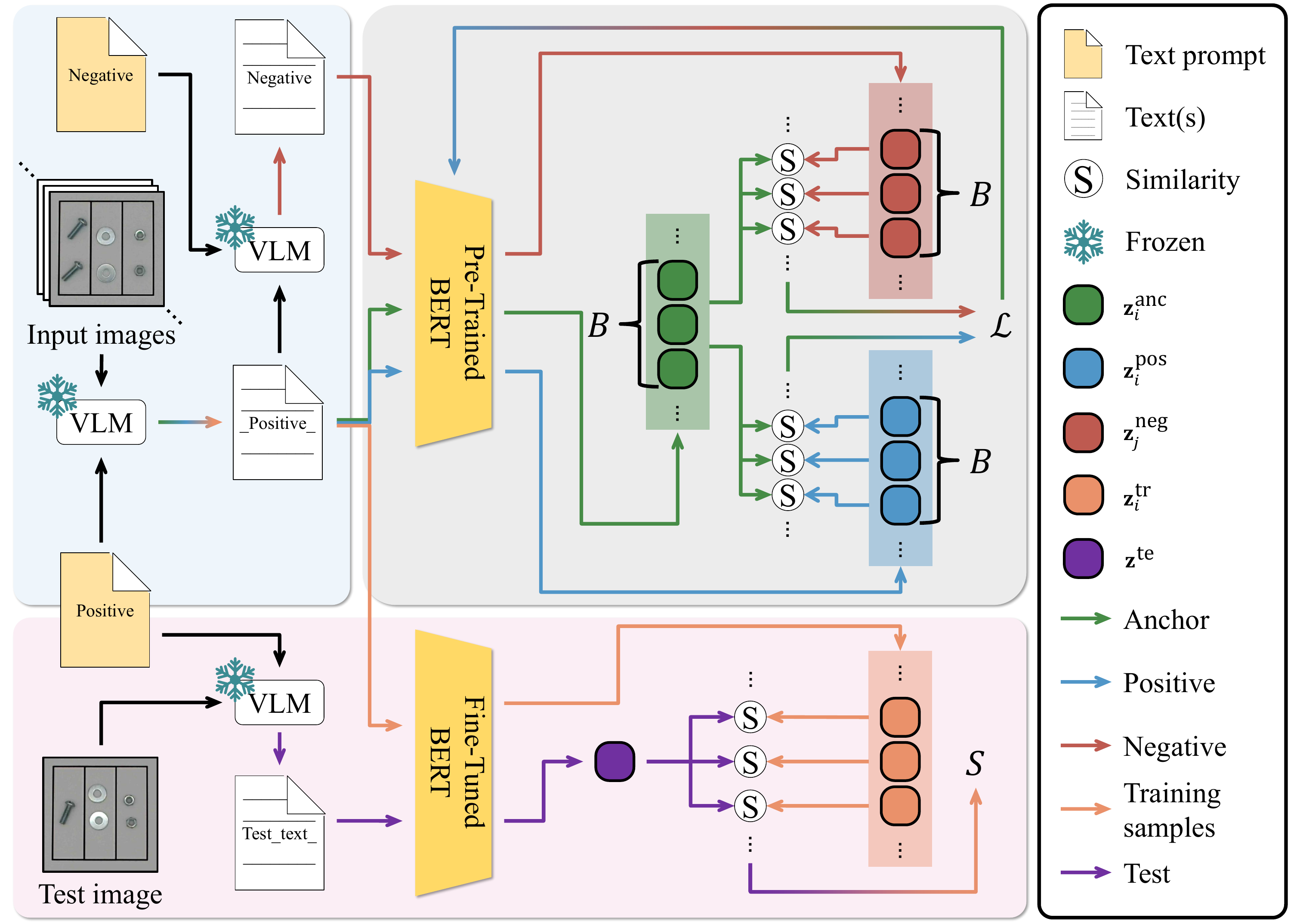}
\caption{
\textbf{Pipeline} of the proposed unsupervised logical anomaly detector.
For each image, a frozen Vision-Language Model (VLM) generates a single logic-focused text conditioned on a scenario-specific text prompt. 
During training, one contradictory negative text is synthesized from each positive text in a text-only manner, and a text encoder (BERT) is fine-tuned via contrastive learning, where dropout-augmented embeddings form the anchor–positive pair and synthesized texts serve as negatives.
During inference, the test text is embedded by the fine-tuned encoder and compared with the set of training embeddings. 
The final normality score is computed by distance-based $k\text{-}$nearest-neighbor aggregation ($k=5$), where larger values indicate more normal samples.
}
\label{Method}
\end{figure*}

\subsection{Problem Setting and Overview}
VID-AD is established as a one-class benchmark for logical anomaly detection, where models learn logical constraints using only normal samples. A primary challenge within this benchmark is the presence of vision-induced distractions, requiring models to decouple underlying logical states from significant appearance variations. Toward this end, we propose a vision-to-text framework that performs detection using semantic descriptions rather than raw images, which ensures that the process prioritizes logical consistency over irrelevant pixel-level fluctuations. Specifically, our approach leverages a frozen Vision-Language Model (VLM) to convert each image into a logic-focused text description guided by scenario-specific prompts (Fig. \ref{Method}). To facilitate training without real anomalies, we first synthesize negative texts by rewriting the normal descriptions into logically inconsistent versions that remain linguistically natural. The model is then trained via contrastive learning to distinguish the original normal texts from their synthesized negative counterparts. During inference, each test image is converted into text through the same VLM process, and the anomaly score is calculated by comparing the text representation with the reference distribution of normal samples via similarity-based aggregation.
This design enables unsupervised detection of logical anomalies while reducing sensitivity to irrelevant appearance variations.

\subsection{Vision-to-Text Description and Negative Synthesis}
This section details the conversion of visual inputs into structured textual descriptions and the subsequent generation of negative training texts.
The conversion process targets specific attributes that define the logical rules of a scenario, including object type, color, count, region, relative length, and spatial relations. By explicitly instructing the frozen VLM to focus on these properties, the system effectively filters out environmental noise such as background texture or lighting conditions. To maintain strict consistency between learning and evaluation, the identical prompt is utilized for both training and testing phases. This consistency is paired with an enforced structured output format to ensure stability in the text embedding space. Such stability is critical under the small-sample regime of VID-AD, where each task typically contains only around 50 images. To align with the specific logic of each task, we tailor the prompt structure to the spatial organization of each scenario, utilizing region-wise descriptions for grid layouts or ordered sequences for relational rules.

\noindent \textbf{Positive text generation (vision-to-text).}
Let $I$ denote an input image in a given task. 
The VLM generates a logic-focused text $d^{pos}$ from $I$ under the text prompt. 
We then apply lightweight text cleanup (e.g., removing system tokens and normalizing whitespace) to obtain a clean text representation. 
Each image yields exactly one such text, which serves as the foundational feature for both contrastive training and anomaly evaluation. This mapping from visual to textual space ensures that the model ignores appearance fluctuations that do not alter the underlying logical state.

\noindent \textbf{Negative text synthesis (text-only rewriting).}
To train the text encoder without anomalous images, we construct negative text examples by substituting key attributes within the original descriptions. 
Specifically, for each $d^{pos}$, we generate a contradictory text $d^{neg}$ that remains fluent and follows the same formatting constraints, while introducing logical inconsistencies. 
To ensure the quality of negative texts, the synthesis process follows three constraints: (i) preserve the text structure and length as much as possible, (ii) apply replacement-only edits (no insertion/deletion that changes the counting semantics), and (iii) enforce at least one attribute-level contradiction (e.g., color, type, count, region, order, relation, or relative length). 
These constraints ensure that the generated negative texts provide challenging training data, which prevents the contrastive objective from being dominated by superficial cues such as sentence length. The resulting training pairs $(d^{pos}, d^{neg})$ allow the model to distinguish logically consistent descriptions from their contradictory counterparts, establishing a robust foundation for unsupervised anomaly scoring in subsequent stages.

\subsection{Contrastive Fine-tuning of Text Encoder}
Building upon the positive-negative textual pairs synthesized in Section 4.2, we fine-tune a text encoder to learn a representation space that captures logical consistency. 
We adopt a pretrained BERT model (bert-base-uncased) \cite{devlin2019bert} and fine-tune it independently for each task under the one-class setting.

\noindent \textbf{Text embedding.}
For an input text $d$, we utilize the BERT model to generate token-level hidden states from its final layer. 
We then compute a comprehensive text embedding through mean pooling across all tokens and subsequent L2 normalization as follows:

\begin{equation}
\label{eq:text-embedding}
\mathbf{h}(d) = \frac{1}{T}\sum_{t=1}^{T}\mathbf{u}_t,\qquad\mathbf{z}(d)=\frac{\mathbf{h}(d)}{\|\mathbf{h}(d)\|_2},
\end{equation}

\noindent where $\mathbf{u}_t$ is the final-layer hidden state of token $t$ and $T$ is the number of tokens.

\noindent \textbf{Contrastive objective with dropout augmentation.}
To establish a highly discriminative embedding space, we train BERT using a contrastive objective that pulls together embeddings of logically consistent descriptions while pushing them away from embeddings of contradictory texts. 
Since VID-AD provides only a single positive description per image, we utilize dropout as a stochastic data augmentation technique to generate positive pairs. Specifically, for each positive text $d_i^{pos}$, we obtain an anchor $\mathbf{z}_i^{anc}$ and a positive embedding $\mathbf{z}_i^{pos}$ by encoding the same input twice. Because dropout randomly deactivates different neurons during each forward pass, these two embeddings represent distinct views of the same semantic content:

\begin{align}
\mathbf{z}_i^{anc} &= \mathrm{Enc}\!\left(d_i^{pos}\right), \\
\mathbf{z}_i^{pos} &= \mathrm{Enc}\!\left(d_i^{pos}\right).
\end{align}

For the corresponding negative text $d_i^{neg}$, we compute $\mathbf{z}_i^{neg} = \mathrm{Enc}\!\left(d_i^{neg}\right)$. 
We maintain the BERT model in training mode during the entire embedding computation process. Consequently, dropout is applied when encoding negative descriptions, while the anchor-positive pair is derived from two independent stochastic realizations of the same positive text. This strategy ensures that the model learns to ignore minor feature variations caused by dropout while remaining highly sensitive to the structural and keyword-based contradictions introduced in the synthesis stage.
Using cosine similarity $\mathrm{sim}(\mathbf{x},\mathbf{y}) = \mathbf{x}^{\top}\mathbf{y}$ between normalized embeddings, we optimize an InfoNCE/NT-Xent style loss \cite{chen2020simple}:

\begin{equation}
\label{eq:contrastive-loss}
\mathcal{L} = -\frac{1}{B}\sum_{i=1}^{B}
\log \frac{\exp(\mathrm{sim}(\mathbf{z}_i^{anc},\mathbf{z}_i^{pos})/\tau)}{Z_i},
\end{equation}

\noindent This loss increases the similarity between the anchor and its matched positive text, while decreasing similarity to contradiction-based negatives.

\begin{equation}
\label{eq:partition-function}
Z_i =
\exp(\mathrm{sim}(\mathbf{z}_i^{anc},\mathbf{z}_i^{pos})/\tau)
+\sum_{j=1}^{B}\exp(\mathrm{sim}(\mathbf{z}_i^{anc},\mathbf{z}_j^{neg})/\tau),
\end{equation}

\noindent where $B$ is the batch size and $\tau$ is a temperature parameter.
Here, $Z_i$ is the normalization term that includes one positive pair and all negative pairs in the batch for anchor $i$.

This training objective maps logically consistent descriptions to a compact region in the embedding space while effectively repelling contradiction-based counterparts.
Consequently, the learned embedding space supports one-class anomaly detection by utilizing the similarity to the training set as a scoring metric in the next section.

\subsection{Inference and Statistical Ensemble Scoring}
During the inference stage, we utilize the same prompt configuration as established in the training phase to generate a descriptive text for each test image through the frozen VLM. The fine-tuned BERT encoder subsequently maps this test description $d^{\text{te}}$ to a normalized embedding $\mathbf{z}^{\text{te}}$ as defined previously. To evaluate the logical consistency of the test sample, we compare this vector against a reference library of training embeddings $\{\mathbf{z}_i^{\mathrm{tr}}\}_{i=1}^{N}$ that are precomputed and stored to represent the normal logical distribution of the task.

\noindent \textbf{Similarity-based anomaly scoring.}
We compute the final normality score through a distance-based $k\text{-}$nearest neighbor rule in the embedding space.
Let $\mathbf{z}^{\mathrm{tr}}_i$ be the embedding of the $i\text{-}$th training text and $\mathbf{z}^{\mathrm{te}}$ be the embedding of a test text, both extracted by the fine-tuned encoder.
The scoring procedure consists of three steps.
First, since all embeddings are $\ell_2$-normalized as defined in Eq. \eqref{eq:text-embedding}, we compute the Euclidean distance from the test text to each training text:

\begin{equation}
d_i=\left\lVert \mathbf{z}^{\mathrm{te}}-\mathbf{z}^{\mathrm{tr}}_i\right\rVert_2.
\end{equation}

Second, we identify the top$\text{-}k$ nearest neighbors:

\begin{equation}
\mathcal{N}_k\!\left(\mathbf{z}^{\mathrm{te}}\right)
= \left\{\, i \;\middle|\; d_i \le d_{(k)}\right\},\quad k=5,
\end{equation}

\noindent where $d_{(k)}$ is the $k$-th smallest value among $\{d_i\}_{i=1}^{N}$ and we set $k=5$ for all experiments. Then, we compute the mean distance over these neighbors:

\begin{equation}
\bar{d}_k=\frac{1}{k}\sum_{i\in \mathcal{N}_k(\mathbf{z}^{\mathrm{te}})} d_i.
\end{equation}

Finally, we convert this mean distance into a bounded normality score:

\begin{equation}
S=\frac{1}{1+\bar{d}_k}.
\end{equation}

A higher score $S$ indicates that the test sample is closer to the training distribution and is therefore more likely to be normal. Conversely, a lower score suggests a significant logical deviation from the established patterns of the training set.

\section{Experiments}
\label{sec:experiments}
\subsection{Experimental Settings}
We conduct an extensive evaluation of our approach and several state-of-the-art vision-based baselines on the VID-AD dataset.
For all vision-based baselines, we follow the respective preprocessing protocols specified by each method and utilize their official implementations with default hyperparameters. For our method, we process images using the default VLM preprocessor \cite{team2024qwen2} and generate a single textual description for each image through a consistent, scenario-specific prompt for training and testing phases.
To ensure reproducibility, we maintain fixed random seeds for both text generation and BERT fine-tuning and will release the prompts, preprocessing scripts, and evaluation code upon the acceptance of this work.

\noindent \textbf{VLM-based text generation.}
We utilize Qwen2-VL-7B-Instruct \cite{team2024qwen2} for image-to-text generation due to its superior balance between description quality and computational efficiency.
To obtain stable and logic-focused descriptions, we generate positive texts using low-temperature decoding with $\gamma=0.001$.
For negative text generation, we use a separate text-only rewriting prompt with explicit contradiction constraints and apply a higher temperature of $\gamma=0.9$ to increase diversity while preserving the original text structure.
Both generation settings are configured with $\mathrm{top}\text{-}p=0.95$ and a maximum limit of 512 tokens.
Specifically, for each positive text, we synthesize exactly one corresponding negative counterpart.

\noindent \textbf{Text encoder training.}
We fine-tune a pretrained BERT model (bert-base-uncased) independently for each task and optimize the contrastive objective in Section 4.3 using Adam \cite{kingma2014adam} with learning rate $2\times 10^{-5}$ and weight decay $10^{-5}$ for $20$ epochs with batch size $B=16$.
The temperature parameter in the NT-Xent loss (Eq. \eqref{eq:contrastive-loss}) is set to $\tau=0.5$, and we apply gradient clipping with a maximum norm of $1.0$ for stability.

\noindent \textbf{Image-level scoring.}
We report image-level anomaly scores for all methods to ensure a consistent evaluation protocol, with performance measured by AUROC (Area Under the Receiver Operating Characteristic curve).
For methods producing pixel-level anomaly maps (PaDiM, PatchCore, EfficientAD, and CSAD), we convert them to an image-level score by taking the maximum response over spatial locations.
For reconstruction-based methods (AnoGAN and VAE), we use the reconstruction error aggregated over pixels as the image-level score.

\subsection{Baselines}
We compare our method with widely used one-class visual anomaly detection baselines that cover feature-space, reconstruction-based, distillation-based, and component-consistency paradigms.
Specifically, we evaluate PaDiM \cite{defard2021padim} and PatchCore \cite{roth2022towards} (feature-space), AnoGAN \cite{schlegl2017unsupervised} and VAE \cite{kingma2013auto, an2015variational} (reconstruction-based), EfficientAD \cite{batzner2024efficientad} (student–teacher distillation), and CSAD \cite{hsieh2024csad} (segmentation-based component consistency).

\subsection{Performance under Different Capture Conditions}

\begin{table*}[ht]
\caption{
\textbf{Condition-wise image-level AUROC} on VID-AD (mean over 10 scenarios for each capture condition). 
BG: background, CD: condition.
Mean$\pm$Std denotes the mean and standard deviation across capture-condition averages.
}
\label{tab:comparison}
\centering
\begin{tabular*}{\hsize}{@{}@{\extracolsep{\fill}}lcccccc@{}}
\toprule
Method & White BG & Cable BG & Mesh BG & Low-light CD & Blurry CD & Mean$\pm$Std \\
\midrule
\makecell[l]{EfficientAD {\cite{batzner2024efficientad}}} & 0.526 & 0.602 & 0.608 & 0.479 & 0.480 & 0.539$\pm$0.057 \\
\makecell[l]{PatchCore {\cite{roth2022towards}}}   & 0.404 & 0.450 & 0.480 & 0.469 & 0.420 & 0.445$\pm$0.029 \\
\makecell[l]{PaDiM {\cite{defard2021padim}}}       & 0.348 & 0.354 & 0.428 & 0.354 & 0.403 & 0.378$\pm$0.032 \\
\makecell[l]{VAE {\cite{an2015variational}}}         & 0.448 & 0.494 & 0.454 & 0.472 & 0.393 & 0.452$\pm$0.034 \\
\makecell[l]{AnoGAN {\cite{schlegl2017unsupervised}}}      & 0.489 & 0.548 & 0.492 & 0.474 & 0.468 & 0.494$\pm$0.028 \\
\makecell[l]{CSAD {\cite{hsieh2024csad}}}        & 0.693 & 0.624 & 0.641 & 0.694 & 0.657 & 0.662$\pm$0.028 \\
\makecell[l]{UniVAD {\cite{gu2025univad}}}        & 0.574 & 0.592 & 0.570 & 0.573 & 0.559 & 0.574$\pm$0.011 \\
Ours & \textbf{0.825} & \textbf{0.811} & \textbf{0.848} & \textbf{0.842} & \textbf{0.826} & \textbf{0.831$\pm$0.013} \\
\bottomrule
\end{tabular*}
\end{table*}

VID-AD enables controlled robustness analysis by preserving the logical state while varying only the capture condition.
As shown in Tab.~\ref{tab:comparison}, our method achieves the best AUROC under all five capture conditions, outperforming the second-best method (CSAD) by 0.132 to 0.207 AUROC across conditions. Furthermore, our approach demonstrates exceptional stability with a standard deviation of 0.013, which is competitive with the most stable baseline (UniVAD) while maintaining significantly higher absolute performance.

The vulnerability of vision-based baselines is particularly evident in the performance gap between their best and worst-performing conditions. While UniVAD maintains a minimal gap of 0.033, this stability is confined to a lower accuracy range, failing to yield competitive results. In contrast, other vision-centric models exhibit significant fluctuations. EfficientAD shows a substantial gap of 0.129 AUROC, while others range from 0.070 to 0.101. Such wide gaps indicate that these methods rely on brittle, low-level visual patterns that are easily disrupted by environmental shifts, suggesting a failure to learn the underlying logical relationships. In contrast, our approach minimizes this fluctuation to a mere 0.037 while maintaining consistently high performance regardless of capture conditions. This stability arises from our transition from pixel-level modeling to language-level reasoning. By converting visual inputs into textual descriptions, our method extracts semantic attributes that remain invariant under low-level visual variations. These results confirm that decoupling logical anomaly detection from raw visual features is effective for achieving robust performance in diverse real-world scenarios.

\subsection{Qualitative Analysis of Vision-centric Evidence}
\begin{figure*}[pos=tb]
\centering
\includegraphics[width=0.8\textwidth]{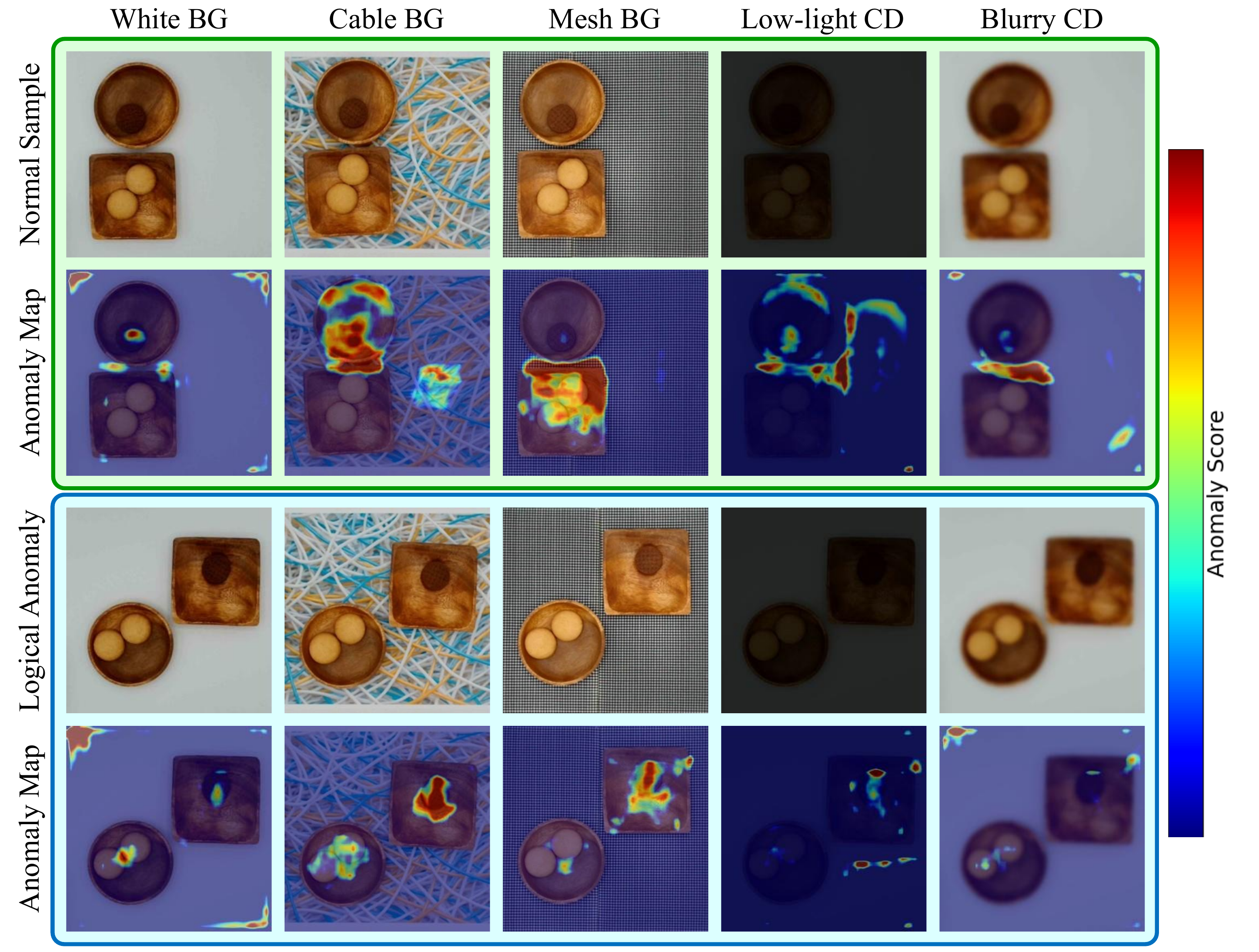}
\caption{
\textbf{Qualitative anomaly maps} of CSAD on the \textit{Cookies} scenario under five capture conditions.
Columns correspond to White BG, Cable BG, Mesh BG, Low-light CD, and Blurry CD.
Rows show, from top to bottom: a normal sample, its anomaly map, a logical anomaly, and its anomaly map.
CSAD's anomaly responses vary substantially with the capture condition, often highlighting background patterns or low-level appearance variations (e.g., repetitive textures in Mesh BG or contrast degradation in Low-light CD) in addition to the relevant objects, illustrating the impact of vision-induced distraction on pixel-level localization.
}
\label{csad_discussion}
\end{figure*}

We qualitatively assess the sensitivity of vision-centric detectors to low-level visual variations. Fig.~\ref{csad_discussion} shows the pixel-level anomaly maps produced by CSAD, the best-performing VAD baseline, for the Cookies scenario under five distinct capture conditions. Despite the identical underlying logical state, CSAD exhibits substantial instability across conditions: textured backgrounds (e.g., Mesh BG) induce spurious activations, whereas capture degradations (Low-light CD and Blurry CD) lead to diffuse or weakened localization. These results suggest that pixel-level evidence in image space can entangle nuisance appearance factors with anomaly cues under vision-induced distraction, thereby yielding anomaly localization that is less reliable and interpretable. 

In contrast, the proposed method avoids pixel-level localization and instead performs detection in the language embedding space, where the scoring depends on the consistency of logic-focused descriptions rather than low-level visual patterns. It is important to note that since our approach is purely language-based, pixel-level anomaly maps are inherently uncomputable, precluding a direct visual comparison in Fig.~\ref{csad_discussion}.
Nevertheless, this decoupling from pixel-level representations is precisely what enables our language-based pipeline to maintain superior stability within the VID-AD setting, as it bypasses the low-level visual distractions that confound traditional vision-centric baselines.

\subsection{Scenario-wise Robustness and Linguistic Expressibility}
\begin{figure}[pos=tb]
\centering
\includegraphics[width=0.5\textwidth]{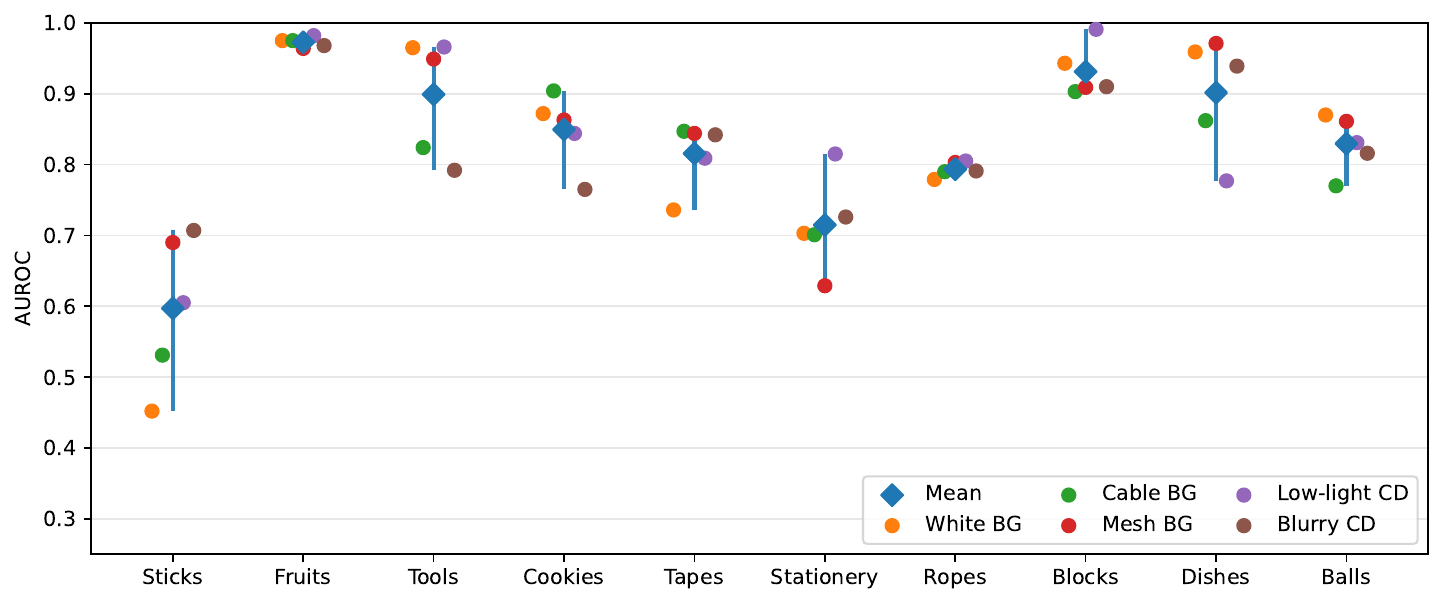}
\caption{
\textbf{Scenario-wise sensitivity} of the proposed method across capture conditions on VID-AD. 
For each scenario, dots show the AUROC under each of the five capture conditions (see legend), the vertical bar indicates the min–max range, and the diamond marker denotes the mean across conditions.
}
\label{condition_sensitivity}
\end{figure}

Condition-wise averages may obscure scenario-dependent robustness differences.
To better reveal this effect, we analyze scenario-wise sensitivity, defined as the standard deviation of AUROC across the five capture conditions for each scenario. The results are summarized in Tab.~\ref{tab:scenario-wise}, with per-condition AUROC distributions further visualized in Fig.~\ref{condition_sensitivity}.
Overall, the sensitivity varies substantially across scenarios, suggesting that robustness under vision-induced distraction is also influenced by the characteristics of the underlying rules.
For our method, sensitivity is particularly low in scenarios such as \textit{Fruits} (0.006) and \textit{Ropes} (0.009). This observation suggests that generated descriptions remain more stable when rule-critical attributes can be expressed as clear and discrete properties. In contrast, scenarios such as \textit{Sticks} (0.096) show higher sensitivity, possibly because the underlying rules are harder to express consistently in a single description when multiple interacting attributes must be captured simultaneously.

To qualitatively support this interpretation, we present VLM-generated positive texts under the five capture conditions for a stable scenario (\textit{Fruits}) and a challenging scenario (\textit{Sticks}) in Tab.~\ref{tab:qual_text_examples}.
The \textit{Fruits} descriptions remain consistent across conditions and preserve rule-relevant attributes. In contrast, \textit{Sticks} descriptions sometimes omit rule-critical relations (e.g., relative length), which leads to incomplete or incorrect rule grounding.
These examples suggest linguistic expressibility may serve as a practical factor related to robustness: language-based detection tends to be more reliable when the scenario admits consistent and complete linguistic grounding of the required logical attributes.
Notably, low sensitivity alone does not guarantee strong detection performance, as a method may consistently yield poor results across conditions.
In contrast, our proposed method achieves both high AUROC (Tab.~\ref{tab:comparison}) and low scenario-wise sensitivity across multiple scenarios (Tab.~\ref{tab:scenario-wise}).
This result highlights the benefit of modeling logical consistency through language representations under controlled appearance shifts.

\begin{table*}[t]
\centering
\caption{
\textbf{Scenario-wise sensitivity} measured as the standard deviation of AUROC across the five capture conditions for each scenario and method.
The smallest standard deviation is highlighted in \textbf{bold}.
}
\label{tab:scenario-wise}
\begin{tabular*}{\hsize}{@{}@{\extracolsep{\fill}}lcccccccccc@{}}
\toprule
Scenario 
& \textit{Sticks} 
& \textit{Fruits} 
& \textit{Tools} 
& \textit{Cookies} 
& \textit{Tapes} 
& \textit{Stationery} 
& \textit{Ropes} 
& \textit{Blocks} 
& \textit{Dishes} 
& \textit{Balls} \\
\midrule
\makecell[l]{EfficientAD {\cite{batzner2024efficientad}}}
& 0.164 & 0.032 & 0.087 & 0.176 & \textbf{0.028} & 0.148 & 0.048 & 0.050 & 0.098 & 0.142 \\
\makecell[l]{PatchCore {\cite{roth2022towards}}}
& 0.062 & 0.115 & 0.128 & 0.098 & 0.167 & 0.133 & 0.063 & 0.159 & 0.022 & 0.152 \\
\makecell[l]{PaDiM {\cite{defard2021padim}}}
& 0.039 & 0.025 & 0.025 & \textbf{0.005} & 0.102 & \textbf{0.058} & 0.047 & 0.072 & 0.065 & 0.106 \\
\makecell[l]{VAE {\cite{an2015variational}}}
& 0.071 & 0.059 & 0.112 & 0.085 & 0.034 & 0.106 & 0.130 & 0.172 & \textbf{0.017} & 0.100 \\
\makecell[l]{AnoGAN {\cite{schlegl2017unsupervised}}}
& 0.149 & 0.050 & 0.116 & 0.111 & 0.077 & 0.171 & 0.110 & 0.049 & 0.074 & 0.172 \\
\makecell[l]{CSAD {\cite{hsieh2024csad}}}
& 0.047 & 0.037 & 0.052 & 0.143 & \textbf{0.028} & 0.232 & 0.142 & 0.090 & 0.038 & 0.228 \\
\makecell[l]{UniVAD {\cite{gu2025univad}}}
& \textbf{0.016} & 0.038 & \textbf{0.028} & 0.080 & 0.039 & 0.115 & 0.032 & 0.121 & 0.105 & 0.093 \\
Ours & 0.096 & \textbf{0.006} & 0.075 & 0.047 & 0.042 & 0.060 & \textbf{0.009} & \textbf{0.033} & 0.073 & \textbf{0.035} \\
\bottomrule
\end{tabular*}
\end{table*}

\begin{table*}[t]
\centering
\caption{
\textbf{Qualitative examples} of VLM-generated positive texts under vision-induced distraction.
We contrast a stable scenario (\textit{Fruits}) and a challenging scenario (\textit{Sticks}), selected based on scenario-wise mean AUROC and condition-wise variability.
For brevity, when the generated text is identical across conditions, we show it once and denote the remaining cells as “Same as White BG.”
Correct indicates whether the generated text preserves the rule-relevant attributes of the scenario, based on our attribute checklist (e.g., quantity/type for \textit{Fruits} and quantity/relative length for \textit{Sticks}).
}
\label{tab:qual_text_examples}
\setlength{\tabcolsep}{3pt}
\renewcommand{\arraystretch}{1.00}
\footnotesize
\begin{tabular}{c >{\centering\arraybackslash}m{1.45cm} >{\centering\arraybackslash}m{1.0cm} >{\raggedright\arraybackslash}m{0.50\linewidth} c}
\toprule
Scenario & Condition & Image & Positive text & Correct \\
\midrule
\multirow[c]{5}{*}{\textit{Fruits}} 
& White BG     & \includegraphics[width=1.0cm]{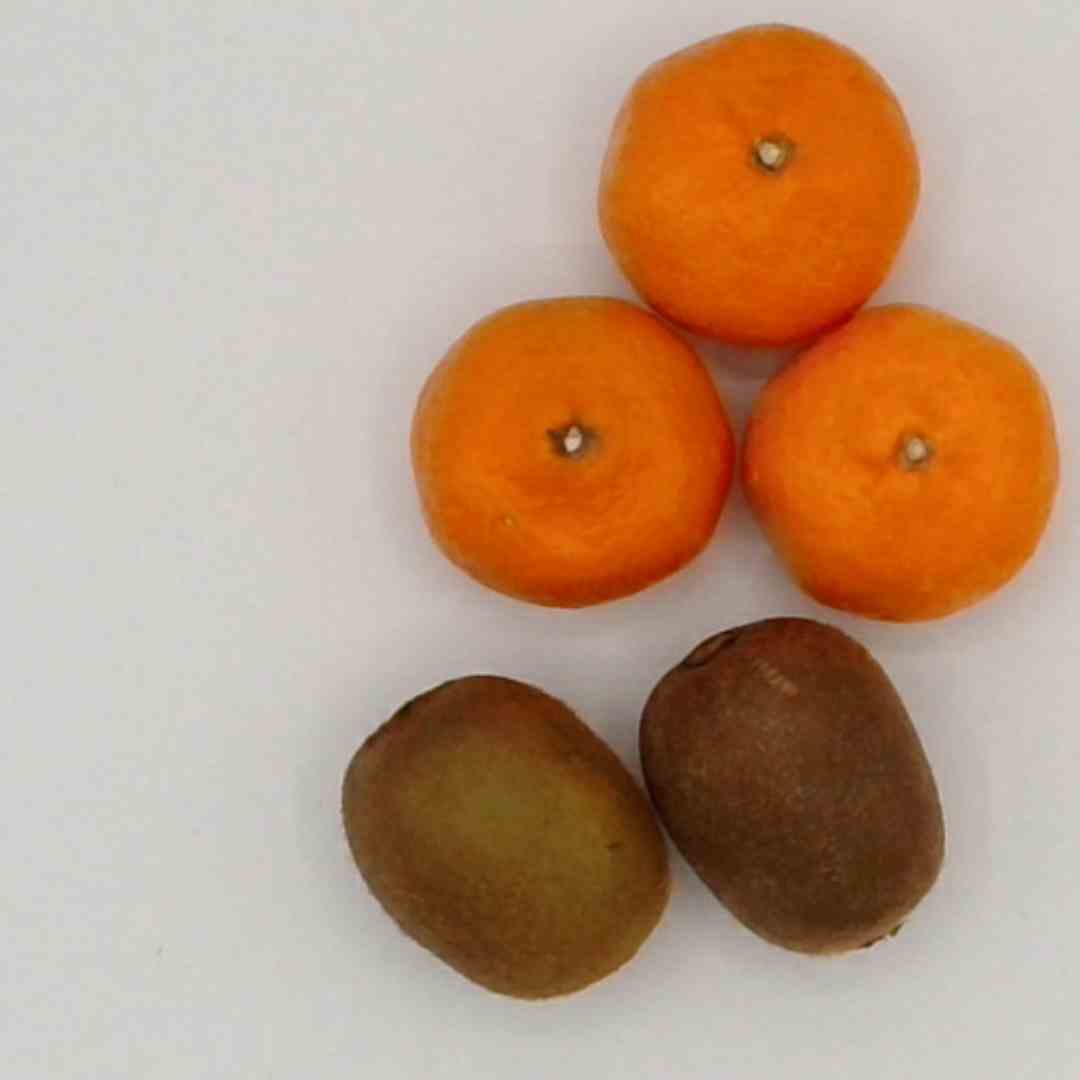} 
& There are three oranges and two kiwis. The total number of items is five. 
& \cmark \\
& Cable BG     & \includegraphics[width=1.0cm]{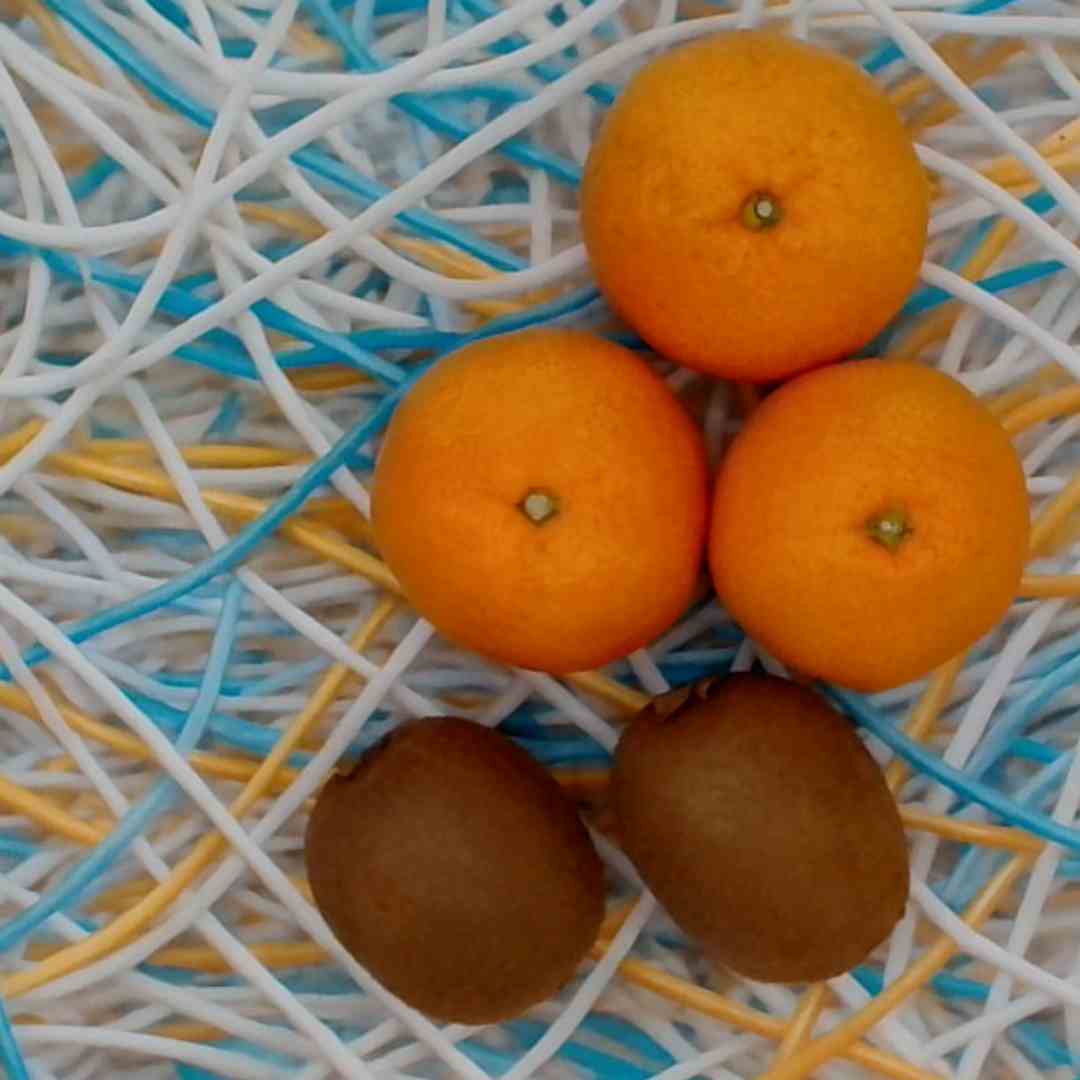} 
& Same as White BG. 
& \cmark \\
& Mesh BG      & \includegraphics[width=1.0cm]{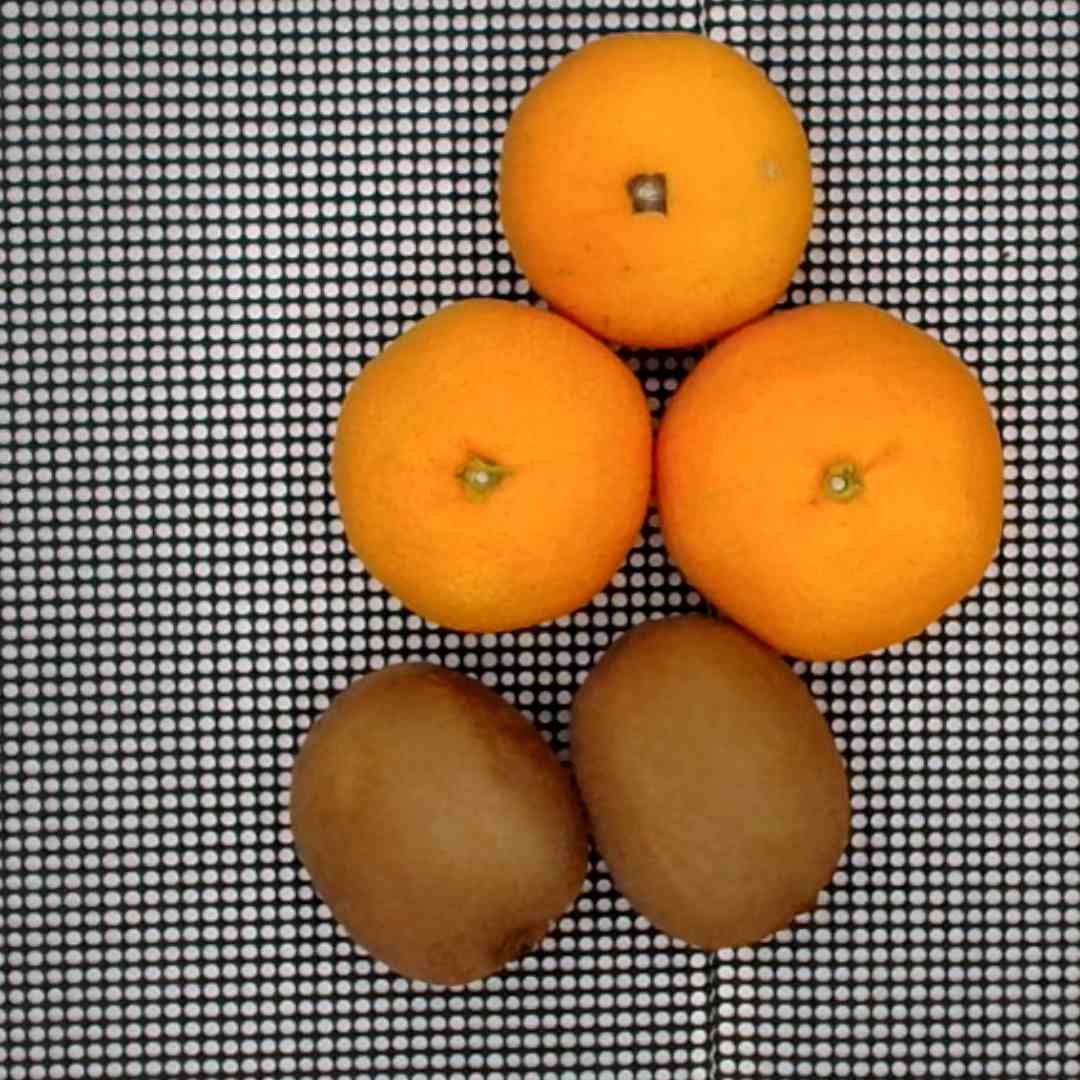} 
& Same as White BG. 
& \cmark \\
& Low-light CD & \includegraphics[width=1.0cm]{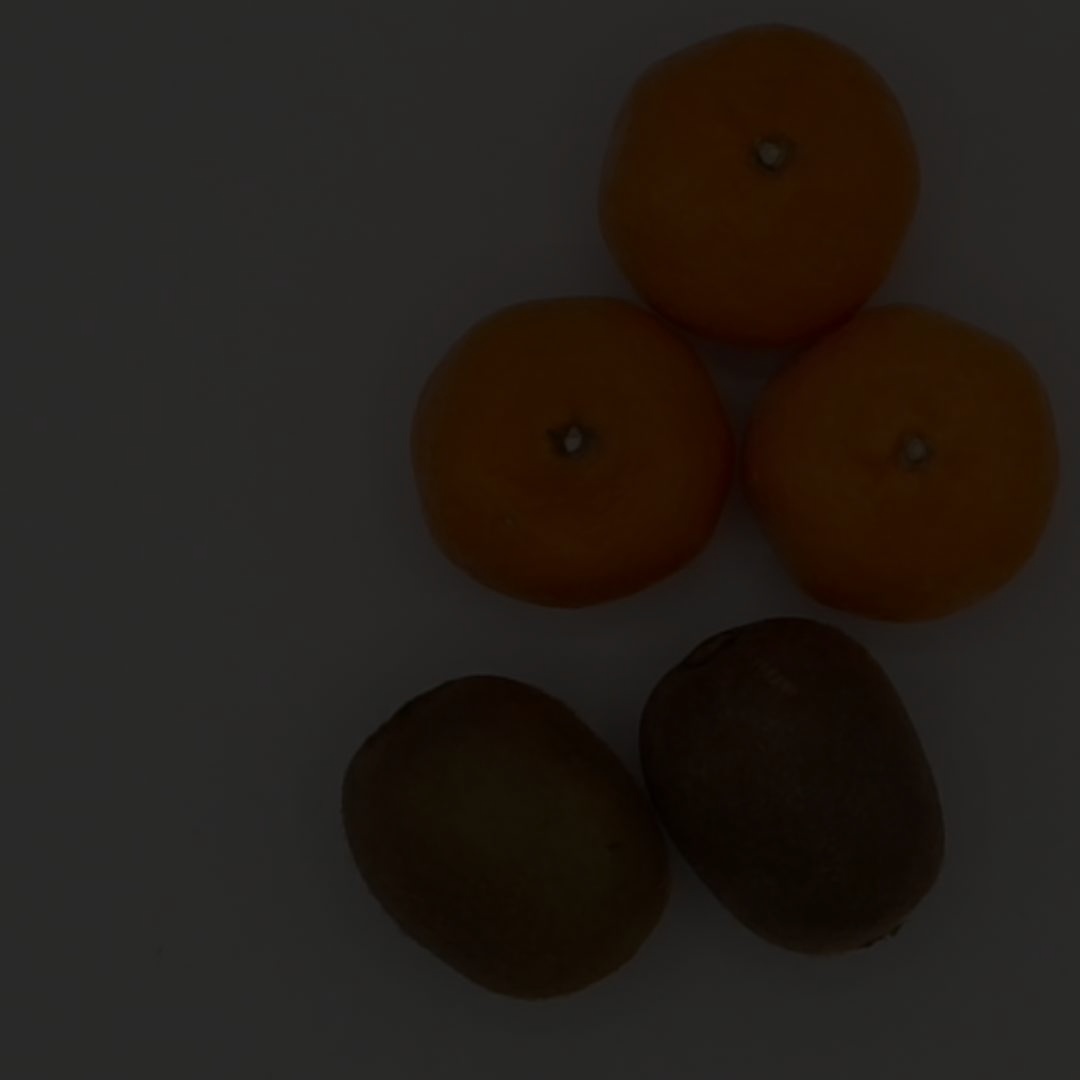} 
& Same as White BG. 
& \cmark \\
& Blurry CD    & \includegraphics[width=1.0cm]{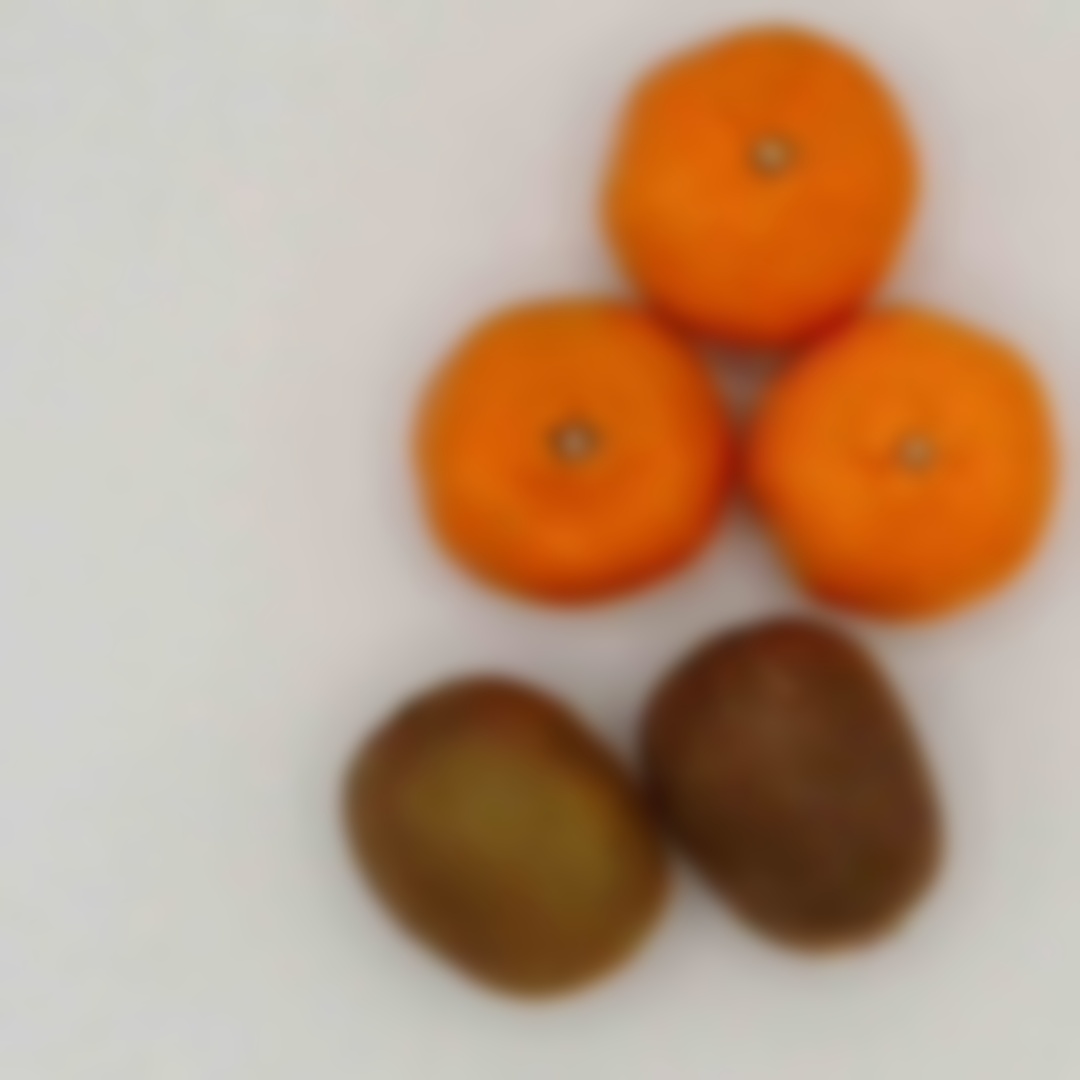} 
& Same as White BG. 
& \cmark \\
\midrule
\multirow[c]{5}{*}{\textit{Sticks}} 
& White BG     & \includegraphics[width=1.0cm]{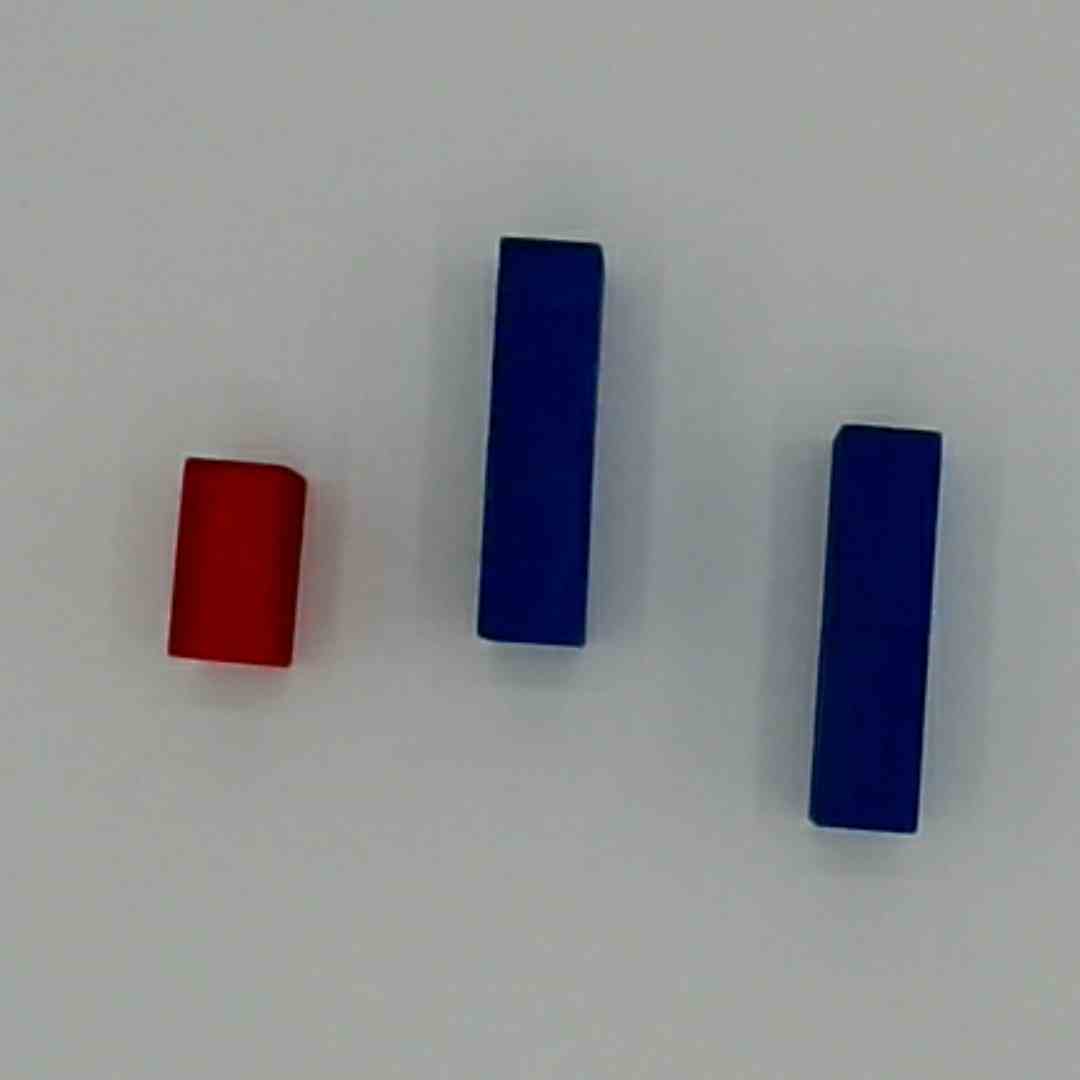} 
& There are three objects in the image. One is a red cylinder, and two are blue rectangles. The shorter blue rectangle is compared to the longer blue rectangle. 
& \xmark \\
& Cable BG     & \includegraphics[width=1.0cm]{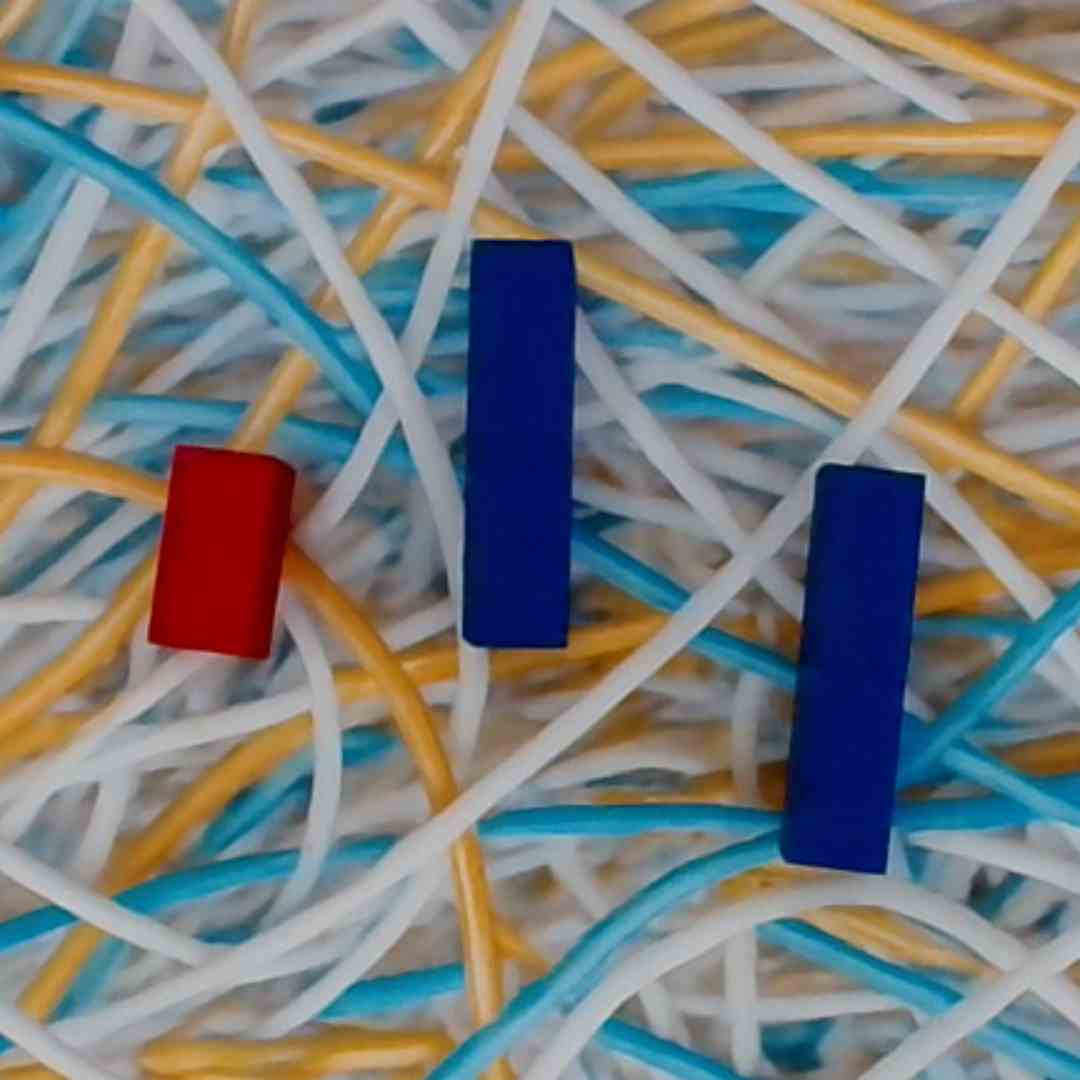} 
& There are three blue rectangular objects, with one being longer than the other two. There is one red rectangular object, which is shorter than the blue objects.
& \xmark \\
& Mesh BG      & \includegraphics[width=1.0cm]{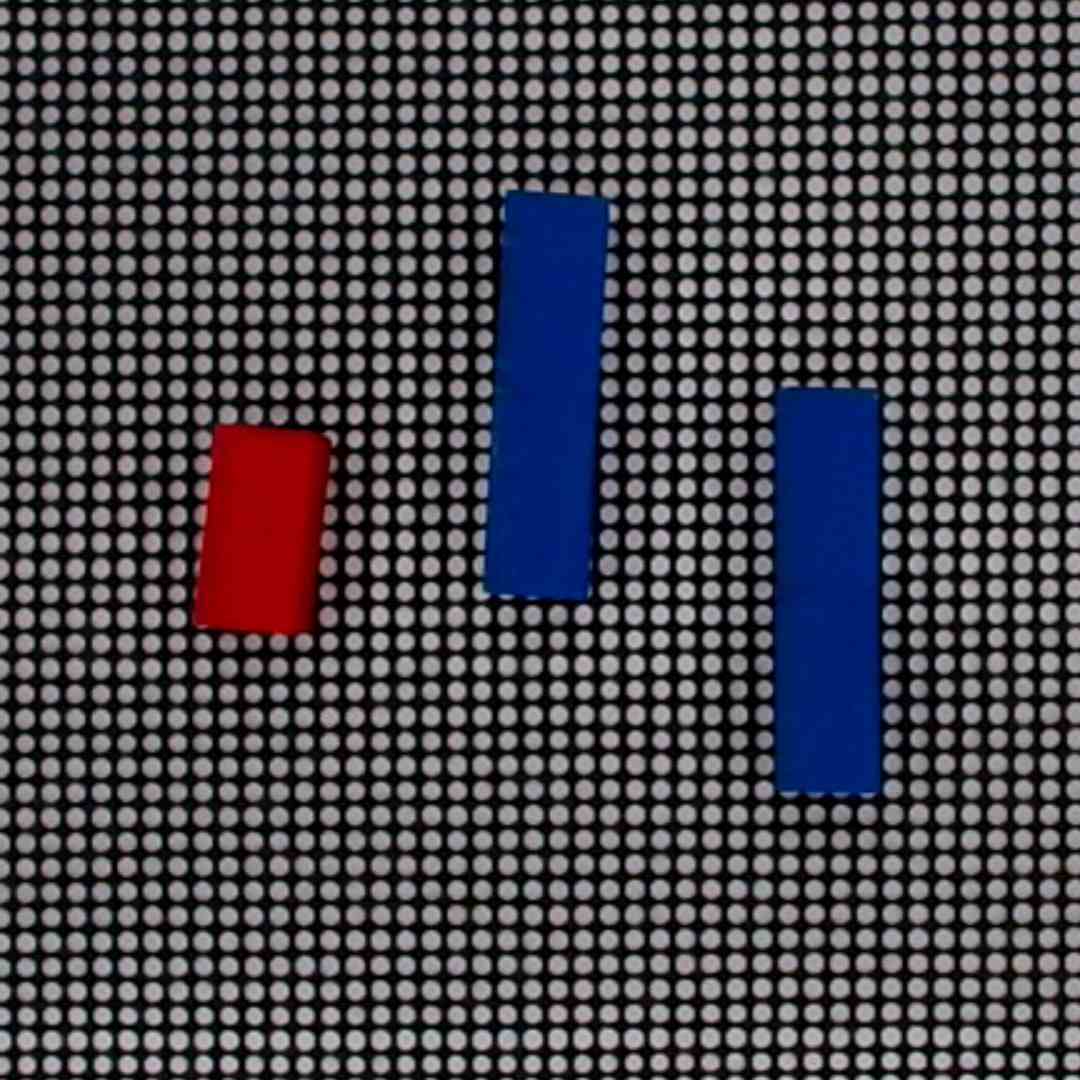} 
& There are three objects in the image: one red object, one shorter blue object, and one longer blue object.
& \xmark \\
& Low-light CD & \includegraphics[width=1.0cm]{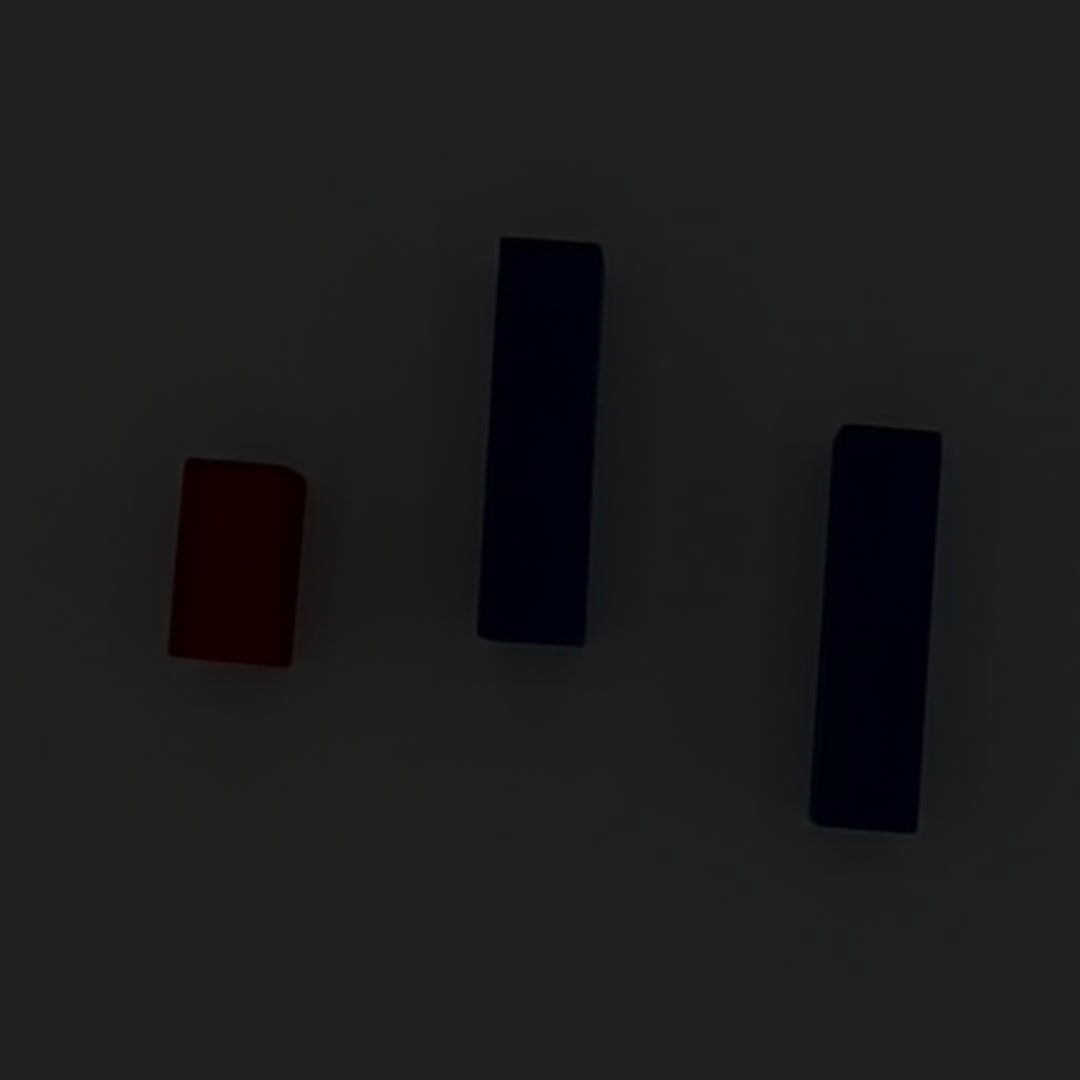} 
& There are three objects in the image. The red object is shorter than the two blue objects.
& \cmark \\
& Blurry CD    & \includegraphics[width=1.0cm]{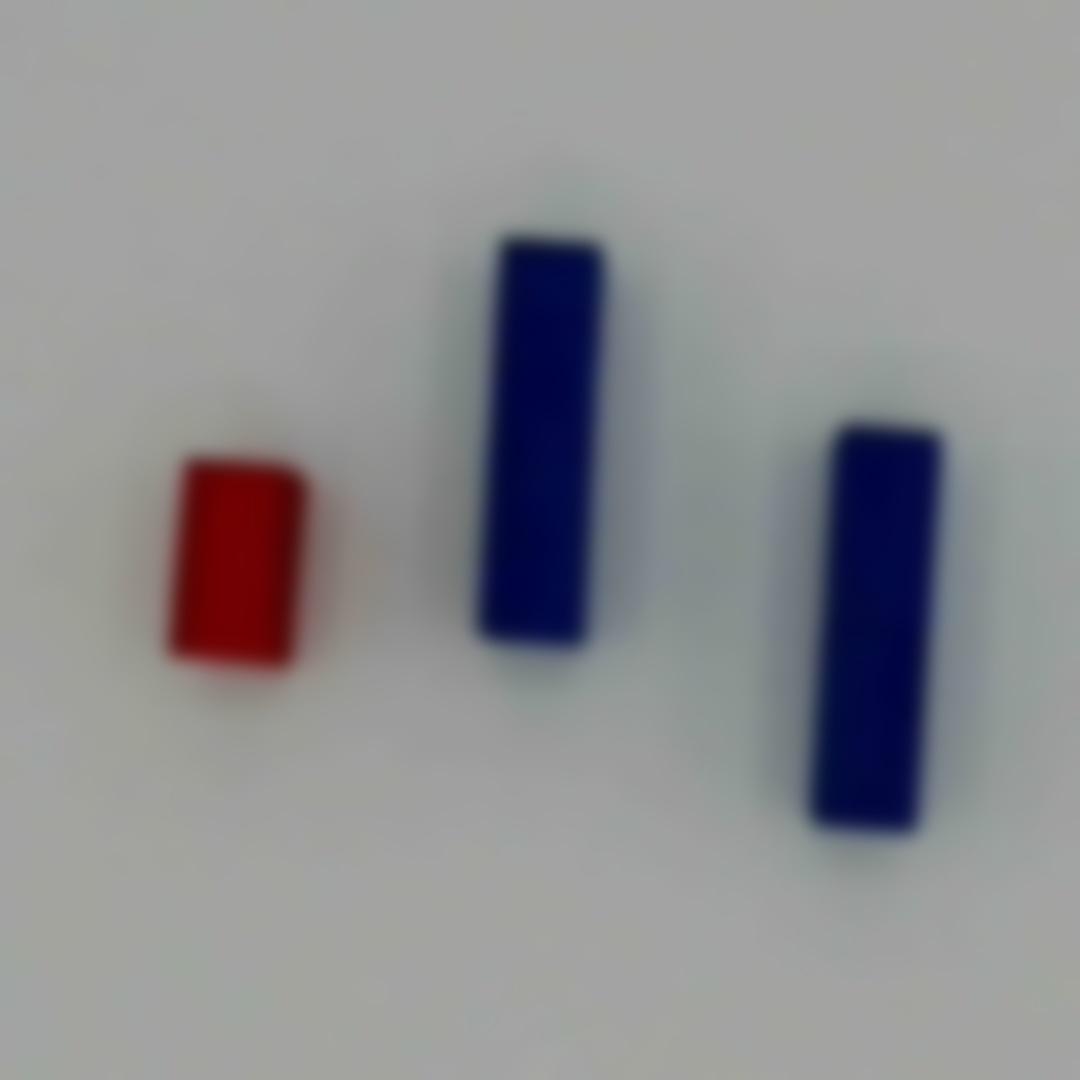} 
& There is one shorter red rectangular object and two longer blue rectangular objects.
& \cmark \\
\bottomrule
\end{tabular}
\end{table*}

\subsection{Ablation Study}

\begin{table*}[ht]
\centering
\caption{
\textbf{Ablation} on the choice of the Vision-Language Model (VLM) used for generating one positive text description per image and one corresponding negative text per positive example. 
For compactness, we use abbreviations in the table: Qwen2 7B = Qwen2-7B-Instruct, Llama3.2 11B = Llama-3.2-11B-Vision-Instruct, and LLaVA 13B = LLaVA-v1.5-13B. 
Results are reported as mean AUROC over the 10 scenarios under each capture condition. 
Mean$\pm$Std denotes the mean and standard deviation across capture-condition averages.
}
\label{tab:ablation}
\begin{tabular*}{\hsize}{@{}@{\extracolsep{\fill}}lcccccc@{}}
\toprule
\textbf{VLM} & White BG & Cable BG & Mesh BG & Low-light CD & Blurry CD & Mean$\pm$Std \\
\midrule
\makecell[l]{Qwen2 7B {\cite{team2024qwen2}}} & 0.825 & 0.811 & 0.848 & \textbf{0.842} & \textbf{0.826} & \textbf{0.831$\pm$0.013} \\
\makecell[l]{Llama3.2 11B {\cite{meta2024llama}}} & \textbf{0.853} & \textbf{0.835} & \textbf{0.861} & 0.769 & 0.793 & 0.822$\pm$0.036 \\
\makecell[l]{LLaVA 13B {\cite{liu2024improved}}} & 0.741 & 0.680 & 0.739 & 0.689 & 0.655 & 0.701$\pm$0.034 \\
\bottomrule
\end{tabular*}
\end{table*}

To isolate the effect of the VLM choice, we compare three Vision-Language Models (VLMs) for vision-to-text conversion: Qwen2-VL-7B-Instruct \cite{team2024qwen2}, Llama-3.2-11B-Vision-Instruct \cite{meta2024llama}, and LLaVA-v1.5-13B \cite{liu2024improved}. The text prompt, BERT encoder, contrastive objective, and scoring strategy are kept the same across all models.
Tab. \ref{tab:ablation} shows that Qwen2 7B achieves the best overall mean AUROC (0.831) with the smallest cross-condition variance (Std = 0.013), followed by Llama3.2 11B (0.822$\pm$0.036) and LLaVA 13B (0.701$\pm$0.034). A closer look at the condition-wise results reveals that Llama3.2 11B attains the highest AUROC under White BG, Cable BG, and Mesh BG. However, its performance degrades substantially under Low-light CD and Blurry CD, increasing variance and reducing the overall mean. In contrast, Qwen2 7B remains strong across both background variations and degraded capture conditions. These results suggest that robust description generation under appearance changes is an important factor for stable logical anomaly detection.

\section{Discussion}
\label{sec:discussion}
\noindent \textbf{Robustness Under Vision-Induced Distraction.}
Our results suggest that robustness under vision-induced distraction is closely linked to representation invariance. Performing detection in a language embedding space that emphasizes logical attributes, rather than raw low-level visual features, reduces sensitivity to nuisance factors such as background changes and capture degradations.

\noindent \textbf{Linguistic Expressibility.}
Across scenarios, the effectiveness of language-based detection depends on linguistic expressibility. The approach appears to be most reliable when rule-critical attributes admit stable and complete linguistic grounding, and tends to degrade when key relational attributes are expressed inconsistently across capture conditions.

\noindent \textbf{Limitations and Future Directions.}
The current framework suggests several avenues for future work.
Although single-text descriptions generated from a fixed prompt provide a controllable and noise-reducing input format, richer or more flexible description strategies could capture finer-grained details that benefit borderline cases.
Reducing prompt dependence, ideally toward prompt-free or minimally prompted generation, would further improve reproducibility and facilitate deployment across diverse manufacturing settings.
More structured textual representations, such as attribute-value tuples or constraint graphs, may also complement the current embedding-based scoring and lead to more explicit modeling of logical rules.



\section{Conclusion}
\label{sec:conclusion}
Logical anomaly detection in industrial inspection remains challenging under low-level visual variations while the underlying logical state remains fixed.
To study this problem, we introduced VID-AD, a one-class benchmark comprising 50 tasks across 10 manufacturing scenarios and five capture conditions.
We also proposed a language-based detection framework that learns from text descriptions rather than visual features. Through contrastive learning between normal descriptions and semantically perturbed descriptions generated via constrained text rewriting, the framework learns textual representations that capture global logical consistency.
Experiments on VID-AD show that representative vision-based baselines degrade under such appearance variations, whereas our method consistently improves performance across all capture conditions, achieving a mean AUROC of 0.831 across all 50 tasks.
Future work will explore reducing prompt dependence to further improve usability and robustness in real deployments.

\section*{CRediT authorship contribution statement}
\textbf{Hiroto Nakata}: Conceptualization, Methodology, Validation, Investigation, Writing - Original Draft. 
\textbf{Yawen Zou}: Writing - Review $\&$ Editing.
\textbf{Shunsuke Sakai}: Writing - Review $\&$ Editing.
\textbf{Shun Maeda}: Writing - Review $\&$ Editing.
\textbf{Chunzhi Gu}: Writing - Review $\&$ Editing.
\textbf{Yijin Wei}: Writing - Review $\&$ Editing.
\textbf{Shangce Gao}: Writing - Review $\&$ Editing.
\textbf{Chao Zhang}: Methodology, Investigation, Resources, Writing - Review $\&$ Editing, Project administration, Funding acquisition, Supervision. 


\section*{Data availability}
The dataset will be publicly released.


\section*{Declaration of competing interest}
The authors have no competing interests to declare that are relevant to the content of this article. All authors certify that they have no affiliations with or involvement in any organization or entity with any financial interest or non-financial interest in the subject matter or materials discussed in this manuscript.





\bibliographystyle{model1-num-names}

\bibliography{triclick-refs}





\end{document}